%% file: wileyNJD-APA.tex
\documentclass[APA,STIX2COL]{WileyNJD-v2}

\usepackage[natbibapa]{apacite}%
\usepackage{balance}
\usepackage{subcaption}

\usepackage[linesnumbered,ruled,algo2e]{algorithm2e}
\usepackage{algpseudocode}
\usepackage{amsmath}
\usepackage{multirow}
\usepackage{makecell}
\usepackage{float}
\usepackage{tabularray}
\usepackage{array,ragged2e}
\usepackage{xcolor}

\SetCommentSty{mycommfont}
\newlength\algowd

\usepackage{textcomp}

\makeatletter
\def\NAT@aysep{,}
\makeatother

\articletype{ORIGINAL ARTICLE}%

\received{NA}
\revised{NA}
\accepted{NA}

\begin{document}

\title{High-Resolution Flood Probability Mapping Using Generative Machine Learning with Large-Scale Synthetic Precipitation and Inundation Data}

\author[1]{Lipai Huang*}
\author[2]{Federico Antolini}
\author[1, 2]{Ali Mostafavi}
\author[2]{Russell Blessing}
\author[3]{Matthew Garcia}
\author[2]{Samuel D. Brody}

\address[1]{\orgdiv{Urban Resilience.AI Lab, Zachry Department of Civil and Environmental Engineering}, \orgname{Texas A\&M University}, \orgaddress{\state{Texas}, \country{USA}}}
\address[2]{\orgdiv{Institute for a Disaster Resilient Texas}, \orgname{Texas A\&M University}, \orgaddress{\state{Texas}, \country{USA}}}
\address[3]{\orgdiv{Civil and Environmental Engineering}, \orgname{Rice University}, \orgaddress{\state{Texas}, \country{USA}}}

\corres{Lipai Huang, Zachry Department of Civil and Environmental Engineering, Texas A\&M University, College Station, TX, USA\\
\email{\href{lipai.huang@tamu.edu}{lipai.huang@tamu.edu}}\\}

\input{abstract}


\maketitle

\input{intro}
\input{method}
\input{results}
\input{discussion}

\input{conclusion}
\section*{Acknowledgments}
The authors gratefully acknowledge the financial support provided by the National Science Foundation under the CRISP 2.0 Type 2 grant (grant number 1832662) and the Texas A\&M X-Grant Presidential Excellence Fund. The perspectives, results, conclusions, and suggestions presented in this research are entirely those of the authors and do not necessarily represent the viewpoints of the funding organizations.

\section*{Code Availability}
The code that supports the findings of this study is available from the corresponding author upon request.

\section*{Data Availability}
The data that support the findings of this study is available from the corresponding author upon request.

\balance
\bibliography{wileyNJD-APA}%

\end{document}

%% file: abstract.tex
\abstract[Abstract]{High-resolution flood probability maps are instrumental for assessing flood risk but are often limited by the availability of historical data. Additionally, producing simulated data needed for creating probabilistic flood maps using physics-based models involves significant computation and time effort, which inhibit its feasibility. To address this gap, this study introduces Precipitation-Flood Depth Generative Pipeline, a novel methodology that leverages generative machine learning to generate large-scale synthetic inundation data to produce probabilistic flood maps. With a focus on Harris County, Texas, Precipitation-Flood Depth Generative Pipeline begins with training a cell-wise depth estimator using a number of precipitation-flood events model with a physics-based model. This cell-wise depth estimator, which emphasizes precipitation-based features, outperforms universal models. Subsequently, the Conditional Generative Adversarial Network (CTGAN) is used to conditionally generate synthetic precipitation point cloud, which are filtered using strategic thresholds to align with realistic precipitation patterns. Hence, a precipitation feature pool is constructed for each cell, enabling strategic sampling and the generation of synthetic precipitation events. After generating 10,000 synthetic events, flood probability maps are created for various inundation depths. Validation using similarity and correlation metrics confirms the accuracy of the synthetic depth distributions. The Precipitation-Flood Depth Generative Pipeline provides a scalable solution to generate synthetic flood depth data needed for high-resolution flood probability maps, which can enhance flood mitigation planning.
}

%% file: intro.tex
\section{INTRODUCTION}\label{sec:1}
Flood hazards pose a significant threat to cities and communities globally, resulting in extensive physical damage and substantial economic costs due to material losses and human casualties, particularly in densely populated regions. In the United States, flood-related insurance claims have consistently averaged around \$40 billion annually over the past four decades \citep{yildirim2022flood}. With the projected increase in the frequency and severity of flood events across various U.S. regions \citep{musselman2018projected, slater2016recent}, this financial burden is expected not only to persist but also to escalate. An accurate assessment of flood risk is crucial to plan for efficient response and recovery of potentially affected communities. Flood risk characterization traditionally involves the creation of floodplain maps, which delineate areas typically prone to flooding based on historical data and hydrologic simulations. However, floodplain maps often have limitations due to their reliance on limited historical data, which may not accurately reflect current and future conditions. Also, development of floodplains using physics-based hydraulic and hydrologic (H\&H) models is computationally expensive. In addition, these maps usually depict flood extent for a few specific flooding frequency, limiting their utility in risk assessments and communication. The gaps in traditional floodplain mapping underscore the need to incorporate probability into flood maps beyond the traditional 100-year and 500-year frequencies. In the context of this study, probabilistic flood maps quantify flood risk from a distribution of rainfall events, without explicitly considering the frequency of each of those events. That is, probabilistic flood maps estimate the likelihood that a rainfall event causes a certain inundation in a given place (e.g., 1 ft) from a large number of rainfall events, without considering the probability of occurrence of each event. For example, if 350 out of 1000 rainfall events cause at least 1 ft inundation in a certain place, a probabilistic flood map will associate a likelihood of 0.35 to the 1-ft inundation of that place for rainfall events with the range of characteristics considered. This quantified likelihood provides a measure of 1-ft inundation risk in the area given the characteristics of rainfall events. In this context, this probability is different from the annual likelihood typically used, which is based on a statistical analysis of historical storms and the extrapolation of flood depth and/or extent values for a few return periods (e.g., 10-, 100-, 500-year storm). By offering a spatially continuous probabilistic measure of flood risks, maps provide quantitative insights that can help policymakers, planners, and communities in the mitigation and preparedness phases with tailored, data-driven strategies that lessen the impacts of potential flooding \citep{adams2022mitigation, nofal2024community}.

The creation of flood probability maps hinges on simulating a large number of flood inundation scenarios based on which the likelihood of a range of flood depths across a region can be estimated. In fact, one of the major barriers to creation of flood probability maps is the lack of numerous flood scenarios. On one hand, the number of historical flood events in a given region is not large enough to build a robust statistical distribution of rainfall events and their associated flood depths. Also, depending on the rate of land use change, historical data may not be adequate to estimate flood risk under current conditions. On the other hand, generating a large number of simulated flood scenarios using H\&H models is computationally expensive and would not be feasible. An alternative to historical data is represented by synthetic high-resolution flood scenarios generated using generative models.  

The use of generative models and machine learning has gained significant attention in flood risk analysis. Recent studies have adopted Generative Adversarial Networks (GANs) to improve data generation, such as \citet{JI2024105896}, which uses GANs to generate synthetic precipitation data integrated with the Soil and Water Assessment Tool (SWAT) to enhance flood frequency analysis. Similarly, \citet{PIADEH2023105772} employs machine learning models within a GAN framework for real-time flood forecasting in urban environments, improving prediction accuracy and emergency response strategies through event-based data modeling. Deep learning models, including Convolutional Neural Networks (CNNs) and Recurrent Neural Networks (RNNs), have been used for predicting flash flood probabilities and integrating with hydrodynamic models for efficient flood mapping and risk assessment \citep{SAVIZNAEINI2024116986}. Machine learning algorithms like Random Forest and Extreme Gradient Boosting (XGBoost) have been applied to assess flood susceptibility and generate flood risk maps by analyzing key flood conditioning factors \citep{ZHU2024101739, LYU2023104744}. Furthermore, studies such as \citet{MCSPADDEN2024100518} and \citet{DOLAGO2023129276} have developed ML-based models, including Long Short-Term Memory (LSTM) networks and conditional GANs, to enhance flood prediction accuracy in specific urban settings. In a related application, \citet{ho2024integrated} employs the Segment Anything model (SAM) combined with vision language models (VLMs) to improve lowest floor elevation (LFE) estimation from street view images, aimed at advancing flood risk assessments. Furthermore, \citet{yin2023integrated} presents an integrated model for assessing the resilience of urban transportation networks under flood disruptions, analyzing traffic flow and network topology to devise strategies for enhancing flood resilience. Despite these advancements, there is currently no model for simulating synthetic flood data specifically for creating probabilistic flood maps.

\begin{figure*}[ht]
    \centering
    \includegraphics[width=0.9\textwidth]{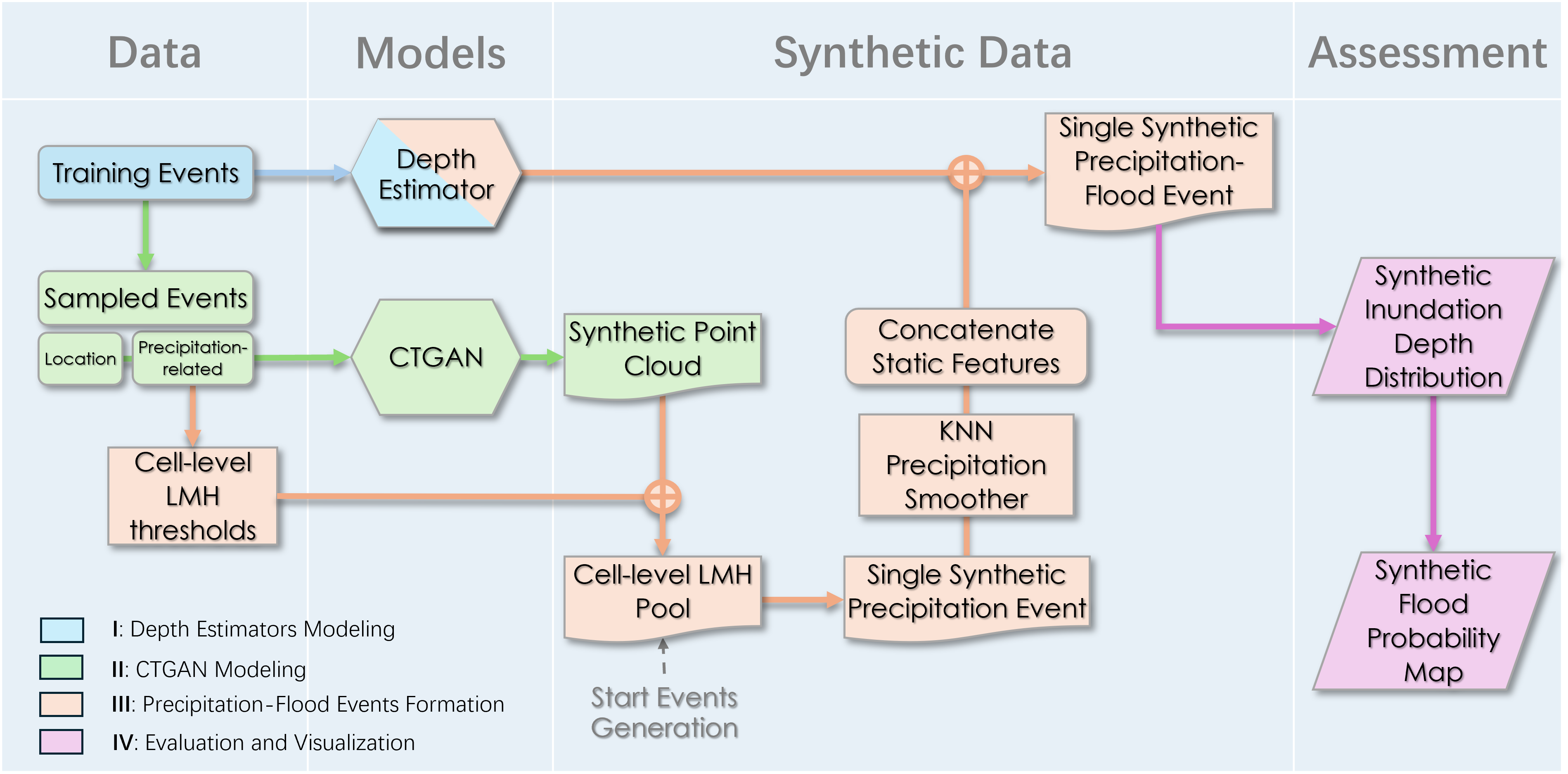}
    \caption{The main research workflow comprises four key steps. \textbf{Step I} involves training and selecting the optimal depth estimation model to generate synthetic flood depth data. The trained model is later applied in \textbf{Step III} for synthetic flood depth construction. \textbf{Step II} focuses on sampling events, preprocessing point-level features, and training a CTGAN with specific constraints to generate a synthetic point cloud through conditional sampling. In \textbf{Step III}, a precipitation feature-level pool is created for each cell mesh using thresholds derived from sampled training event features, enabling strategic sampling to generate synthetic precipitation event features. The trained depth estimator from Step I then processes KNN-smoothed synthetic precipitation features along with static features to produce synthetic flood depths. Finally, \textbf{Step IV} iterates \textbf{Step III} to generate thousands of synthetic precipitation events, forming a synthetic depth distribution and ultimately constructing a synthetic flood probability map.}
    \label{fig:main}
\end{figure*}

To address these challenges, we propose the Precipitation-Flood Depth Generative Pipeline, a novel methodology that harnesses generative machine learning to synthesize large-scale precipitation data and predict corresponding flood inundation depths through strategic sampling and modeling. The main workflow is illustrated in Fig. \ref{fig:main}. Precipitation-Flood Depth Generative Pipeline starts with the training of a depth estimator using a number of H\&H model-generated flood events, incorporating precipitation-based, spatial, and region-specific features. Following this, a Conditional GAN (CTGAN) with constraints is employed to generate synthetic rainfall precipitation events. Strategic thresholds are established to filter these synthetic records, ensuring close alignment with true precipitation patterns. For each location, synthetic events are smoothed using a K-nearest neighbors (KNN) algorithm and processed through the trained depth estimator to derive synthetic depth distributions. By iterating this procedure across multiple synthetic rainfall events, we construct flood probability maps for different combinations of rainfall characteristics (duration, peak precipitation intensity, and cumulative precipitation). These maps are validated using similarity and correlation metrics to confirm the correspondence of synthetic depth distributions to training data.

The novel contributions of the model presented in this paper are threefold:
\begin{itemize}
    \item [$\bullet$] We introduce a novel methodology that utilizes a surrogate machine learning pipeline to estimate flood probabilities using synthetic precipitation-flood events generated by CTGAN and depth estimator.
    \item [$\bullet$] The cell-wise machine learning modeling, introduced in Section \ref{sec:depth_estimator}, presents an innovative approach to feature engineering, yielding enhanced prediction performance compared to global models.
    \item [$\bullet$] We develop an all-to-one event sampling algorithm designed to strategically improve the quality of synthetic records while preserving non-linearity, ensuring a more realistic simulation of flood events.
\end{itemize}

\textit{Flood-Precip GAN} provides a scalable solution for generating high-resolution flood probability maps, to support to flood preparedness and mitigation efforts. The following section explains the components of the \textit{Flood-Precip GAN} model and the associated datasets.

%% file: method.tex
\section{MATERIAL AND METHODS}\label{sec:2}
\subsection{Study Region and Data}
\subsubsection{Study Region}
\begin{figure}[ht]
    \centering
    \includegraphics[width=0.5\textwidth]{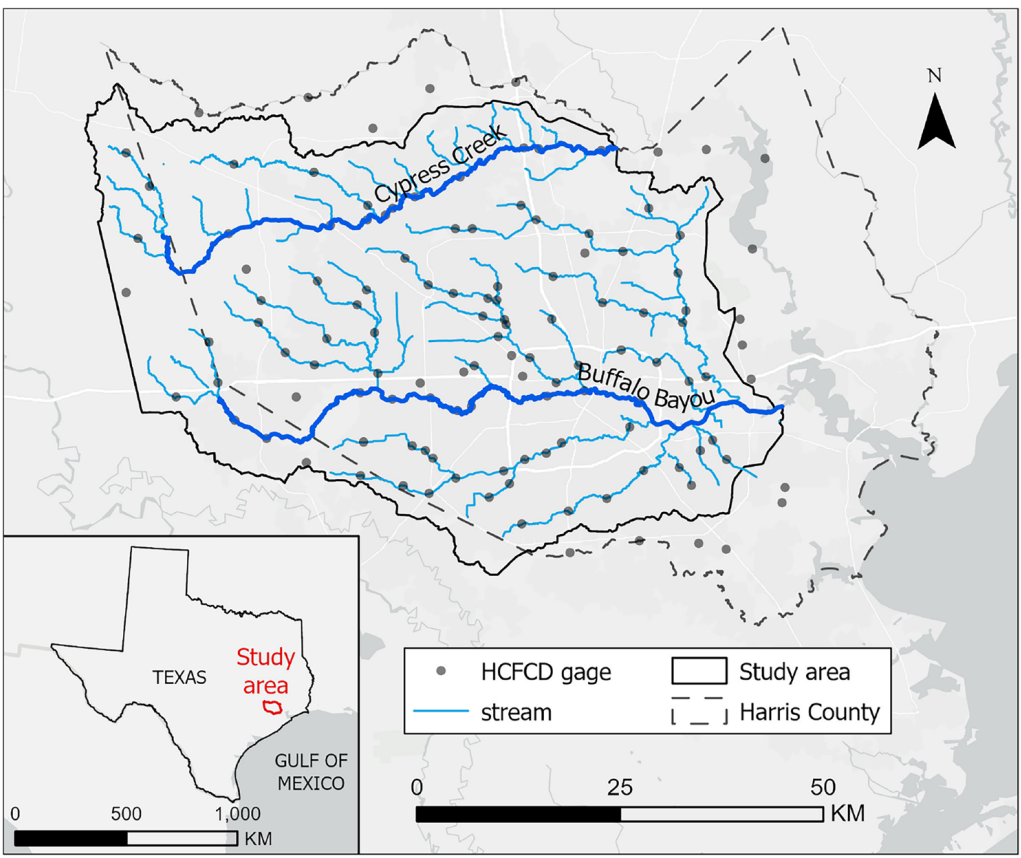}
    \caption{Study area and Harris County Flood Control District gauge distribution. Map generated using \textit{ArcGIS Pro 3.0.0} (\url{https://pro.arcgis.com/}).}
    \label{fig:study_area}
\end{figure}
As shown in Fig. \ref{fig:study_area}, Harris County, Texas, serves as the test bed for developing the Precipitation-Flood Depth Generative Pipeline. The primary watersheds in this area are Cypress Creek and Buffalo Bayou. Harris County, located at the heart of the Greater Houston Metropolitan Statistical Area, covers 1,778 square miles (4,605 square kilometers) and has seen its population grow to over 4.5 million in the past decade. The county's flat topography, with elevations ranging from -40 feet (-12.19 meters) to 300 feet (91.44 meters), coupled with its largely developed land (over two-thirds), along with 20\% pasture and cultivated lands \citep{dewitzNationalLandCover2021}, contributes to significant flood risk. The Cypress Creek watershed in the northern part and the Buffalo Bayou system in the central and southern areas flow into the San Jacinto River and the Ship Channel, eventually draining into the Gulf of Mexico. Dense urban development, poor natural drainage, limited soil infiltration, and the county's subtropical climate further exacerbate its vulnerability to chronic flooding.

\subsubsection{Data}
In creating Precipitation-Flood Depth Generative Pipeline, there is a need for an initial set of rainfall events and related flood inundation scenarios. The lack of sufficient historical precipitation flood inundation data was addressed using 592 precipitation events processed with a physical-based model \citep{garciaphdthesis2023} to obtain corresponding flood scenarios. Specifically, \cite{garciaphdthesis2023} considered storm events occurred in Harris County in the period 2014-2022, and applied Rasterized Time-series Resampling Method (RTRM) to build 592 synthetic, realistic storms varying by spatial extent, duration, and rainfall intensity. \cite{garciaLeveragingMeshModularization2023} modeled those events in HEC-RAS 2D and produced as many flood scenarios. The study region in HEC-RAS 2D was modeled using a 1,200 ft $\times$ 1,200 ft mesh grid, with additional refinements applied near major rivers and tributaries to enhance both computational accuracy and stability. This final mesh consists of 26,301 individual cells. Manning’s roughness coefficient and imperviousness values were sourced from the 2019 National Land Cover Dataset \citep{dewitzNationalLandCover2021}, while soil infiltration values were obtained from the Gridded Soil Survey Geographic database \citep{usdanrcsGriddedSoilSurvey2023}. Model validation involved analyzing data from Harris County Flood Control District gauges \citep{harriscountyfloodcontroldistrictHarrisCountyFlood2023} during five significant flood events from 2016 to 2020, including the 2016 Tax Day flood and Hurricane Harvey in 2017. This validation process, which calibrated the HEC-RAS 2D model for global minimum error across all gauges, confirmed the model's reliability and accuracy for simulating diverse flood scenarios \citep{garciaLeveragingMeshModularization2023}. We used the 592 rainfall events as the initial set from which to generate further synthetic rainfall events, as described in Section \ref{sec:ctgan} and Section \ref{sec:single_event}. The flood depth scenarios became the input for the depth estimator, a surrogate machine learning model for determining cell-wise maximum flood depth, described in Section \ref{sec:depth_estimator}. 

\subsection{Preprocessing}\label{sec:preprocess}
We employed Voronoi tessellation \citep{markus_konrad_2021_4531339} to partition the training dataset, assigning each point within the study area a unique polygonal region to ensure there is no overlap among cells. Additionally, we segmented the 592 rainfall events used in the HEC-RAS 2D flood scenarios into three classes by inundation depth level, to ensure consistent stratification for subsequent dataset splitting and resampling. Specifically, for the training set, events are classified with mean depths of $\leq 2\ \text{inches}$, between $> 2$ and $\leq 6\ \text{inches}$, and $> 6\ \text{inches}$ in a ratio of 2:4:1, respectively. The validation set consists of 20\% of the training set volume, with the remaining data designated as the test set. This approach optimizes the separation of events to ensure relatively balanced representation across the training, validation, and testing phases, facilitating more effective modeling.

\subsection{Depth Estimator Model}\label{sec:depth_estimator}
For the depth estimator component of Precipitation-Flood Depth Generative Pipeline, we chose a tree-based machine learning model trained on simulated scenarios from the HEC-RAS 2D model. Compared to traditional machine learning models, tree-based models have shown superior performance due to their ability to manage complex, non-linear relationships and interactions within the data. XGBoost \citep{chen2016xgboost, shehadeh2021machine}, in particular, excels with its gradient boosting framework, which iteratively improves model accuracy by minimizing residual errors from previous iterations. Additionally, deep learning models in regression, such as the Transformer \citep{vaswani2017attention, castangia2023transformer}, utilize attention mechanisms to capture long-range dependencies and complex feature interactions, providing substantial benefits in predictive performance. In this study, we employed the XGBoost regressor and the regression Transformer as universal depth estimators, training them on physics-based model-generated events that include both spatial features and precipitation-based features, as shown in Table \ref{tab:features}. This integration aims to enhance the accuracy and robustness of maximum flood inundation estimations. Furthermore, we compared these universal models with a cell-wise model, MaxFloodCast V2 \citep{lee2024predicting}, an XGBoost-based depth estimator aggregating 26,301 machine learning models, one for each of the 26,301 cells of the mesh. In MaxFloodCast V2 each cell model is trained independently of the others, without considering precipitation or water levels in neighboring cells. However, spatial dependencies exist in the natural water flow. For example, the water level of cells located in a channel is influenced by the water draining from upstream areas. To account for this spatial influence while maintaining cell-wise model independence, we introduced a heavy cumulative precipitation ratio (HCPR) and a heavy peak precipitation ratio (HPPR). Heavy precipitation is the hourly peak intensity or cumulative rain volume above a certain threshold established for a study area. HCPR and HPPR are defined at the watershed level as in Eq. \ref{eq:1} and Eq. \ref{eq:2}. 

\begin{equation}
\text{R}_i^c = \frac{\sum_{j}^{}AC_{i,j}\times h^c_{i,j}}{AW_{i}}, c \in \{\mathfrak{c}, \mathfrak{p}\}
\label{eq:1}
\end{equation}

\begin{equation}
    h_{i, j}^c =
        \begin{cases}
            1, & \text{if } p_{i, j}^c > 2 \text{ in}, \\
            0, & \text{otherwise},
        \end{cases}
    \label{eq:2}
\end{equation}
where $c$ represents HCPR as $\mathfrak{c}$ and HPPR as $\mathfrak{p}$, with $R$ denoting the respective heavy precipitation ratio. $AW_i$ refers to the total area of watershed $i$, while $AC_{i,j}$ represents the area of cell $j$ within watershed $i$. The binary variable $h_{i,j}$ identifies whether cell $j$ in watershed $i$ experiences heavy precipitation, and $p_{i,j}$ denotes the corresponding precipitation condition for cell $j$ within the same watershed. HCPR and HPPR are defined at the watershed scale, and are uniform across all cells within the same watershed during a rainfall event. These ratios can differ considerably across watersheds and across events, and offer substantial explanatory potential as surrogates for precipitation patterns.
\begin{table*}[]
    \centering
    \caption{Features used in Depth Estimator.}
    \label{tab:features}
    \begin{tabular*}{\textwidth}{@{\extracolsep\fill}p{0.3\textwidth}p{0.1\textwidth}p{0.18\textwidth}p{0.34\textwidth}@{}}
        \toprule
        \textbf{Feature} & \textbf{Level} & \textbf{Unit} & \textbf{Description} \\
        \midrule
        Cumulative Precipitation & All & Inch & Event total precipitation \\
        Peak Precipitation & All & Inch & Event peak precipitation \\
        Duration & All & Hour & Effective precipitation duration \\
        Channel & Universal & Binary (1 or 0) & Identification of channel cell \\
        Terrain Elevation & Universal & Feet & Geographical height of each cell \\
        9 HCPRs (one for each watershed) & Cell-wise & Scalar (range 0 to 1) & Proportion of watershed area experiencing heavy cumulative precipitation \\
        9 HPPRs (one for each watershed) & Cell-wise & Scalar (range 0 to 1) & Proportion of watershed area experiencing heavy peak precipitation \\  
        \bottomrule
    \end{tabular*}
\end{table*}

We assessed the predictive performance of the depth estimators (in Section \ref{sec:estimator_select}) using Root Mean Squared Error (RMSE) and R-squared ($R^2$) scores \citep{chicco2021coefficient} to evaluate both accuracy and model explainability. RMSE measures the average magnitude of the prediction errors, providing insight into the estimator's precision, while $R^2$ indicates the proportion of variance in the observed data that is predictable from the input features, reflecting the model’s explanatory power. In addition, we conducted a comprehensive analysis of the aggregated performance of these models on both channel and non-channel cells. This distinction is crucial due to the varying hydrological and ground texture characteristics inherent to these environments, which influence the uncertainty and behavior of flood dynamics.

\subsection{Rainfall Data Augmentation via CTGAN}\label{sec:ctgan}
Given the limited physics-based model simulated precipitation-flood events, to generate relatively robust synthetic rainfall events in our study, we strategically sampled 315 events as explained in Section \ref{sec:preprocess} and then employed Conditional Generative Adversarial Networks (CTGAN) from SDV \citep{SDV, ctgan} for tabular data augmentation. CTGAN extends the GAN framework to handle structured data, particularly tabular datasets, making it highly effective for generating synthetic data in scenarios with small, imbalanced datasets. One of CTGAN's strengths is its ability to support the specification of constraints and conditions during the data generation process, enabling the preservation of specific relationships or attributes present in the original dataset.

We converted the training cell-level data into point-level data using centroids to facilitate generation. Consequently, the input attributes for CTGAN included latitude, longitude, cumulative precipitation, peak precipitation, and precipitation duration. To enhance the quality of the synthetic records, we constrained conditional sampling to ensure that synthetic points satisfied the following conditions: (1) sampled points are inside the study area (Fig. \ref{fig:study_area}); (2) for a sampled point the cumulative precipitation is always greater than the peak precipitation value; (3) for a sampled point the peak precipitation is greater than the average precipitation intensity, calculated by dividing the cumulative precipitation by the event duration.

To determine the optimal hyperparameters for the generator and discriminator learning rates, as well as the stoppage epochs, we conducted a comprehensive hyperparameter grid search to explore the optimal learning rates for the generator and discriminator, ensuring both networks improved steadily without overpowering each other. Additionally, selecting appropriate stoppage epochs enabled us to terminate the training process at the most accurate point, preventing overfitting or underfitting. The optimal checkpoint was selected based on the corresponding synthetic dataset achieving the highest average marginal distribution, as evaluated using the Kolmogorov-Smirnov statistic (KS statistic) \citep{monge2023two}. The KS statistic $K$ for a particular feature $x_i$ between the training dataset and the synthetic dataset is defined as follows:
\begin{equation}
    K_{i,n,m} = \sup_{x_i}|F_{t,n}(x_i)-F_{s,m}(x_i)|
    \label{eq:ks}
\end{equation}
Where $F_{t,n}$ and $F_{s,m}$ represent the empirical distribution functions of the training dataset and the synthetic dataset generated by CTGAN, respectively. The function $\sup_{x_i}$ denotes the supremum across the domain of feature $x_i$. The overall quality score for all features, excluding location-based features, in the synthetic dataset is computed using the following formula:
\begin{equation}
    Score = \frac{\sum^N_{i=1}(1-K_{i,n,m})}{N}
    \label{eq:score}
\end{equation}
Where $N$ is the number of features. A higher score indicates a higher quality of the synthetic data, signifying a closer approximation to the training data distribution. The outcome of this step is a comprehensive synthetic point cloud containing 10,000,000 synthetic rainfall records generated by CTGAN, representing various precipitation durations. This diverse point cloud presents a significant challenge in processing and formulating individual precipitation events under varying conditions.
\subsection{Rainfall Events Generation}\label{sec:single_event}
In the synthetic point dataset, which comprises collections of precipitation records, we have predefined thresholds for cumulative precipitation, peak intensity, and duration respectively—categorized as Low, Medium, and High (LMH)—for each cell mesh based on training data. These categories are established based on the conditions shown on table \ref{tab:thresh}.
\begin{table}[] 
    \centering
    \caption{Definition of LMH Thresholds for Each Precipitation-Based Feature within a Cell Mesh.}
    \label{tab:thresh}
    \begin{tabular*}{\linewidth}{@{\extracolsep{\fill}}lp{0.5\linewidth}@{}}
        \toprule
        \textbf{Class} & \textbf{Range} \\
        \midrule
        Low    & $[0\ ,\ \mu_i - \theta_1\cdot\sigma_i]$ \\
        Medium & $(\mu_i - \theta_1\cdot\sigma_i\ ,\ \mu_i + \theta_2\cdot\sigma_i]$ \\
        High   & $(\mu_i + \theta_2\cdot\sigma_i\ ,\ +\infty)$ \\
        \bottomrule
    \end{tabular*}
\end{table}

Where $\mu_i$ is the mean of a precipitation-based feature $x_i$ from training dataset, and $\sigma_i$ refers to its standard deviation. $\theta_1$ and $\theta_2$ are two positive constants. Consequently, for each cell mesh, we defined low (L), medium (M), and high (H) thresholds based on the distribution of precipitation-related features—cumulative precipitation, peak precipitation, and duration—derived from the training dataset. These thresholds were then applied to the synthetic point dataset generated as described in Section \ref{sec:ctgan}. For each cell, we constructed a corresponding precipitation feature-level pool comprising all possible LMH combinations of cumulative precipitation, peak precipitation, and duration, resulting in a total of 27 combinations (i.e., LLL, LLM, ..., HHH.). To ensure consistency and representativeness, we systematically iterated over all synthetic point records, retaining indices of synthetic points that fell within the cell boundary while preserving the original distribution of the pool.

Establishing the synthetic indices pool for each cell enabled to generate global precipitation events with various precipitation distributions across the study area. To generate an event a specific duration was sampled randomly from the aggregated pool and set as the global rainfall duration, reducing the possible cumulative-peak-duration combinations from 27 to 9 (e.g., if the input synthetic duration is H, then the potential set of combination is {LLH, LMH, LHH, MLH, MMH, MHH, HLH, HMH, HHH}). Subsequently, a synthetic record was randomly sampled  for each cell from its own pool. The sampling was repeated for all the cells of the mesh. This strategic sampling allowed to create single synthetic precipitation events from a large point cloud, transitioning from point-level to cell-level representation in accord with the original distribution of precipitation characteristics in the study area. To enhance the approximation of training precipitation events, we applied the KNN method as a smoother \citep{wagner2017k} to cumulative and peak precipitation features respectively. Additionally, we computed HCPR and HPPR from these smoothed values. Following this cell-level processing, we utilized the depth estimator with optimal predictive performance as identified in Section \ref{sec:depth_estimator} to generate inundation depths resulting from each synthetic precipitation event. The depth estimator was applied to the thousands of generated rainfall events and produced a synthetic inundation depth distribution in each cell. 

enabling the estimation of the frequency of rainfall events that cause certain flood depth levels in different areas in producing the synthetic flood probability maps.

To assess the event-generation performance, we evaluated the statistical similarity between the inundation depth distribution of the training set of rainfall events and the inundation depth distribution of the synthetic set. We implemented interpolation using the formula in Eq. \ref{eq:interpolate} for the synthetic set depth distribution $\mathbf{d_s}$ to ensure the same dimension as the training set depth distribution $\mathbf{d_t}$, facilitating the evaluation. Let $\mathbf{d_t} = [d_t(1), d_t(2), \ldots, d_t(m)]$ be the training set depth vector with $m$ records, and $\mathbf{d_s} = [d_s(1), d_s(2), \ldots, d_s(n)]$ be the synthetic set depth vector with $n$ synthetic records. We define an interpolation function $f$ such that:

\[
f(i) = d_s(i) \quad \text{for} \quad i = 1, 2, \ldots, n
\]

The interpolated values $d_s'(j)$ for $j = 1, 2, \ldots, m$ are given by:

\begin{equation}
\label{eq:interpolate}
d_s'(j) = f\left(\frac{j-1}{m-1} \cdot (n-1) + 1\right)
\end{equation}

This process ensures that $\mathbf{d_t}$ and $\mathbf{d_s'}$ are of the same length, enabling valid comparisons using multiple statistical measures. Cosine similarity \citep{vijaymeena2016survey} assesses their directional similarity. Pearson correlation \citep{freedman2007statistics} evaluates the linear relationship between the two distributions. Finally, Kullback-Leibler (KL) divergence \citep{hershey2007approximating} quantifies how one probability distribution diverges from the other. The corresponding formulas can be written as:
\begin{equation}
\text{Cosine Similarity}(\mathbf{d_t}, \mathbf{d_s'}) = \frac{\mathbf{d_t} \cdot \mathbf{d_s'}}{\|\mathbf{d_t}\| \|\mathbf{d_s'}\|}
\end{equation}
\begin{equation}
\text{Corr}(\mathbf{d_t}, \mathbf{d_s'}) = \frac{\sum_{i=1}^{m} (d_t(i) - \bar{d_t})(d_s'(i) - \bar{d_s'})}{\sqrt{\sum_{i=1}^{m} (d_t(i) - \bar{d_t})^2 \sum_{i=1}^{m} (d_s'(i) - \bar{d_s'})^2}}
\end{equation}
\begin{align}
D_{KL}(\mathbf{p} \parallel \mathbf{q}) &= \sum_{i=1}^{m} p(i) \log \frac{p(i)}{q(i)}, \\
p(i) &= \frac{d_t(i)}{\sum^m_{k=1} d_t(k)}, \label{eq:9}\\
q(i) &= \frac{d_s'(i)}{\sum^m_{k=1} d_s'(k)} \label{eq:10}
\end{align}

where $\mathbf{p}$ and $\mathbf{q}$ are the normalized versions of $\mathbf{d_t}$ and $\mathbf{d_s'}$, respectively, as represented in Eqs. \ref{eq:9} and \ref{eq:10}. Besides the averaged synthetic depth distribution across the study region, we further explore the performance by aggregating the results into channel and non-channel levels, as described in Section \ref{sec:depth_estimator}. This detailed analysis allows to assess how well the synthetic data captures the unique characteristics of different regions in the study area.

%% file: results.tex
\section{RESULTS}\label{sec:3}
\subsection{Selection of Depth Estimator}\label{sec:estimator_select}
To understand the advancements of tree-based models and transformers in regression problems, specifically in the context of inundation depth estimation, we compared the predictive performance of two universal models, XGBoost Regressor and Regression Transformer, with one cell-wise model, MaxFloodCast V2. MaxFloodCast V2 utilizes an XGBoost-based architecture and incorporates heavy precipitation ratio features. We trained the Regression Transformer using an encoder with 4 layers, a model dimensionality of 128, a feed-forward layer size of 2,048, and 8 attention heads with a 25\% dropout rate on 8 NVIDIA RTX A100 GPUs. The XGBoost Regressor was trained on an NVIDIA RTX A6000, while MaxFloodCast V2 was trained in parallel on an AMD EPYC 7702P 64-Core Processor, sharing the same hyperparameter settings as the XGBoost Regressor. The model's objective function aimed to minimize RMSE, utilizing a learning rate of 0.01. The XGBoost algorithm generated 1000 trees with a maximum depth of 5, applying L1 regularization to mitigate overfitting and employing a subsample ratio of 0.3 to introduce non-linearity and improve model robustness. Following the sampling strategy described in Section \ref{sec:preprocess}, we configured 252 precipitation-flood events for training, 63 events for validation, and the remaining events as the test set. The comparison results are presented in Table \ref{tab:estimator_compr}, which assesses the overall, channel, and non-channel RMSE, as well as the $R^2$ score.
\begin{table*}[]
    \centering
    \caption{Depth Estimator Comparison on test dataset.}
    \label{tab:estimator_compr}
    \begin{tabular}{|c|c|c|c|c|c|c|c|c|}
        \hline
        \textbf{Model} & \textbf{Level} & \makecell{\textbf{Overal}\\ \textit{RMSE}} & \makecell{\textbf{Overal}\\$R^2$} & \makecell{\textbf{Channel}\\ \textit{RMSE}} & \makecell{\textbf{Channel}\\$R^2$} & \makecell{\textbf{Non-channel}\\ \textit{RMSE}} & \makecell{\textbf{Non-channel}\\$R^2$} & \textbf{Time}\\
        \hline
        \makecell{Regression Transformer\\ \citep{vaswani2017attention}} & Universal & 2.6634 & 0.6537 & 4.1264 & 0.7278 & 2.4133 & 0.4928 & 5.8s\\
        \hline
        \makecell{XGBoost Regressor\\ \citep{chen2016xgboost}} & Universal & 2.6916 & 0.6463 & 4.2755 & 0.7078 & 2.4153 & 0.4920 & 4.1s\\
        \hline
        \makecell{MaxFloodCast V2\\ \citep{lee2024predicting}} & Cell-wise & 0.6996 & 0.9189 & 1.7280 & 0.9107 & 0.5680 & 0.9200 & 10.6s\\
        \hline
    \end{tabular}
\end{table*}

The comparison of results reveals significant differences in the predictive performance between the universal models and the cell-wise model. The Regression Transformer and XGBoost Regressor both show substantial errors, with overall RMSE scores above 2 ft, indicating they struggle to accurately predict inundation depths. Their performance is particularly poor in channel cells, where RMSE values exceed 4 ft. In contrast, MaxFloodCast V2 demonstrates superior performance with an overall RMSE of 0.6996 ft and an $R^2$ value of 0.9189, significantly outperforming the universal models. It achieves lower RMSE in both channel (1.7280 ft) and non-channel cells (0.5680 ft), highlighting its accuracy and reliability. The cell-wise approach of MaxFloodCast V2, which tailors the model to the specific characteristics of each cell, allows it to better capture local variations in precipitation and terrain features, resulting in more accurate predictions. Also, the incorporation of heavy precipitation ratio features enhances its ability to capture the influence of runoff from upstream areas on channel cell depths. The dataset used for training, generated by the physics-based model, is both limited and imbalanced, comprising 90 traing events. This limitation affects the universal models more significantly, as they are less adept at handling such imbalances compared to the specialized cell-wise approach of MaxFloodCast V2. Overall, while the cell-wise model demands higher computational resources and longer processing times than universal models, its advantages outweigh these costs in the context of this pipeline, which focuses on region-specific flood depth estimation. Given its superior performance in capturing complex hydrological dynamics, the computational overhead is a minor trade-off, making MaxFloodCast V2 the optimal depth estimator for the Precipitation-Flood Depth Generative Pipeline.

\begin{figure*}[ht]
    \centering
    \begin{subfigure}[b]{0.8\textwidth}
        \centering
        \includegraphics[width=\textwidth]{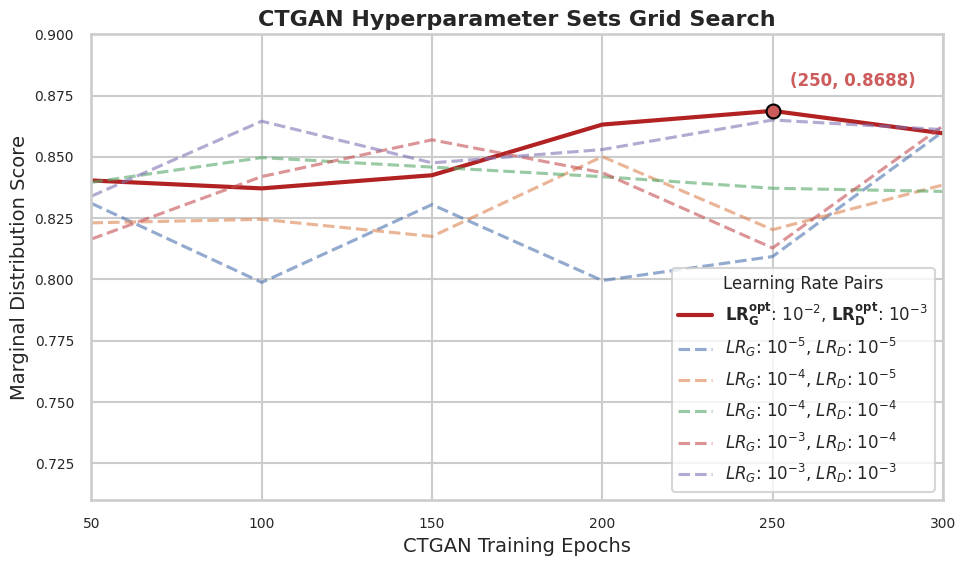}
        \captionsetup{font=normalsize}
        \caption{}
        \label{fig:grid_search}
    \end{subfigure}
    
    \vspace{0.5cm} 

    \begin{subfigure}[b]{\textwidth}
        \centering
        \includegraphics[width=\textwidth]{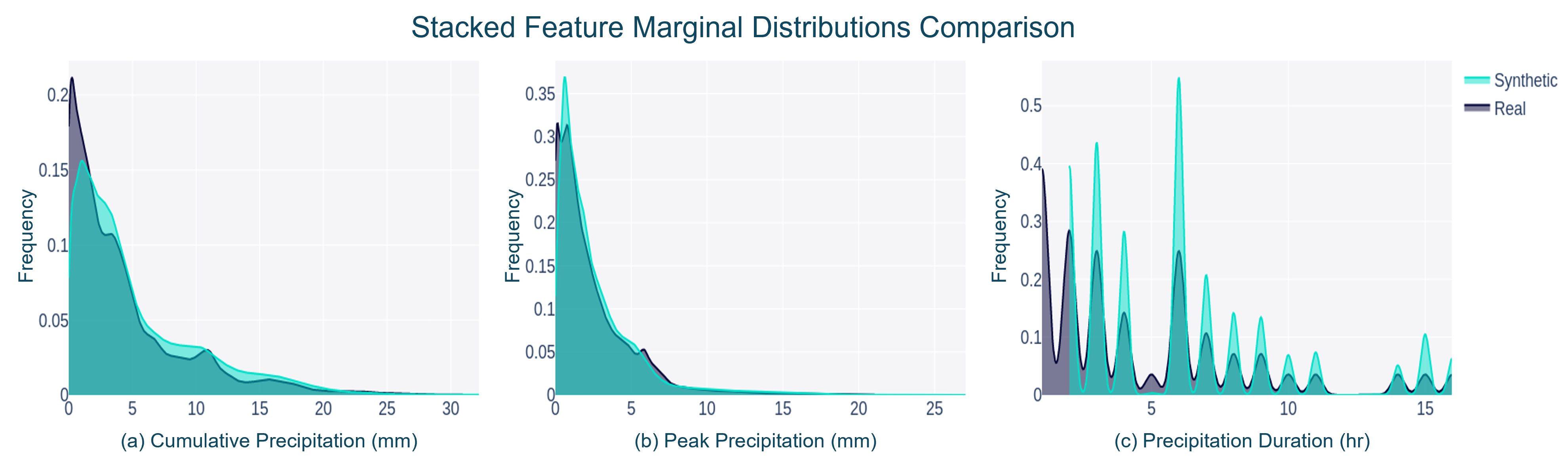}
        \captionsetup{font=normalsize}
        \caption{}
        \label{fig:real_vs_syn}
    \end{subfigure}
    
    \caption{(I) Grid search results for CTGAN hyperparameters. The optimal hyperparameter set was selected based on the highest average marginal distribution scores across all synthetic precipitation-based features generated by the best-performing CTGAN configurations. The key hyperparameters explored in the grid search include the generator and discriminator learning rates, as well as the number of training epochs. (II) Stacked Marginal Distribution Comparison of Three Synthetic Features between training data and synthetic data: (a) cumulative precipitation, (b) peak precipitation and (c) duration. The distribution of the training dataset is represented in gray, while the synthetic distribution is depicted in light blue. Distributions were generated using the Synthetic Data Vault (SDV) \citep{SDV}.}
\end{figure*}

\subsection{CTGAN-generated Precipitation Records Quality Review}\label{sec:res_ctgab}
Among the initial 592 precipitation-flood events, we strategically sampled 90 events following the 2:4:1 ratio described in Section \ref{sec:preprocess} and configured 36 hyperparameter sets for our grid search. This involved combining six pairs of learning rate settings with six distinct stoppage epochs, increasing from 50 epochs to 300 epochs in increments of 50, and modeling on NVIDIA's RTX A6000 GPU. This meticulous grid search focused on generating records under the condition $\text{duration} \geq 1$, targeting the potential simulation of global precipitation events. As shown in Fig. \ref{fig:grid_search}, we identified the optimal checkpoint with a generator learning rate of $10^{-2}$, a discriminator learning rate of $10^{-3}$, and early stopping at 250 epochs. Notably, the generator learning rate is set to be 10 times higher than that of the discriminator to prevent the generator from being overpowered in the early stages before it can effectively learn the underlying data patterns. Ultimately, both the generator and discriminator converged before reaching the 300-epoch limit. The optimal settings enabled a direct comparison of cumulative precipitation, peak precipitation, and duration between the training dataset and the 10,000,000 synthetic data points generated by CTGAN, as depicted in Fig. \ref{fig:real_vs_syn}. This configuration produced an average marginal distribution score of 0.802, demonstrating a high fidelity of the synthetic data, particularly for cumulative and peak precipitation when compared to the original dataset. As shown in Fig. \ref{fig:real_vs_syn}(c), we enforced a constraint in CTGAN that limits $\text{duration} \geq 1$, since our focus is primarily on global precipitation event generation. Additionally, the CTGAN model maintains the flexibility to generate local precipitation events by allowing cumulative precipitation to drop to zero. Moving forward, the challenge remains to construct multiple precipitation-depth events from the synthetic point dataset that closely correlate with training events.

\subsection{Synthetic Rainfall Event Assessment}\label{sec:synth_rainfall}
Using the strategic filtering and sampling methods outlined in Section \ref{sec:single_event}, we generated 10,000 synthetic rainfall events through parallel computing. For each event, the sampled global duration level guided and constrained the selection of precipitation features from each cell’s tailored pool. Leveraging the 27-element precipitation level combination pool described in Section \ref{sec:single_event}, the cell-level sampling preserved regular precipitation trends while maintaining the probability of extreme scenarios. As the number of simulated events increased, the overall synthetic precipitation patterns closely aligned with those in the training dataset. Due to the independence and nonlinearity of point data in CTGAN-generated records, cumulative and peak precipitation maps often exhibited a "pepper-salt" appearance. To mitigate this, we applied KNN smoothing, selecting the value of $K$ based on the synthetic event’s duration following that the longer duration correspond to larger $K$ values, and vice versus. From these smoothed precipitation-based features, we derived HCPR and HPPR, resulting in 21 input features for the depth estimator, MaxFloodCast V2, which was then used to generate synthetic inundation depths and complete the precipitation-flood event. The event generation process averaged 10 minutes for 100 rainfall events in parallel, utilizing 100 GB of memory on an AMD EPYC 7702P 64-core processor.

\begin{figure*}[]
    \centering
    \includegraphics[width=0.75\textwidth]{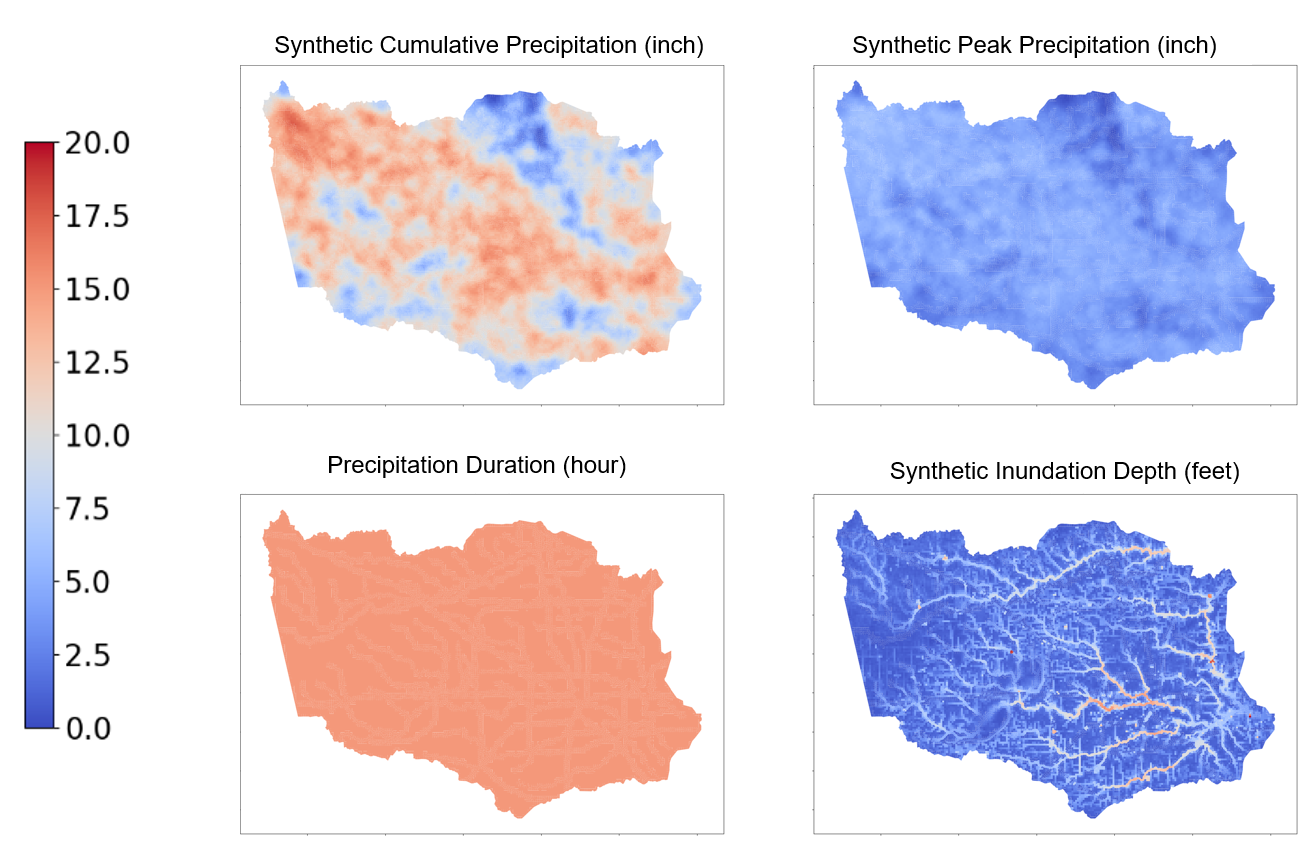}
    \caption{Synthetic flood event with 15 hours global precipitation. All the attribute maps are in the same scale and they share the same color bar with different units: inch, inch, hour and feet respectively. The synthetic cumulative precipitation and peak precipitation are processed by 50-NN smoother. Maps generated by \textit{Geopandas} Python package.}
    \label{fig:syn_event_15}
\end{figure*}

Fig. \ref{fig:syn_event_15} illustrates a synthetic precipitation-flood event with a global duration of 15 hours, comprising four subplots: synthetic cumulative precipitation, synthetic peak precipitation, precipitation duration, and synthetic inundation depth. In Harris County this 15-hour storm event would have a return period of approximately 200 years \cite{perica2018precipitation}. Cumulative and peak precipitation maps show some spatial variability, as expected for a relatively long but still sub-daily rainfall. The inundation depth map highlights flood-prone regions, particularly along river channels and low-lying areas. This result is consistent with the expectation that a 200-year storm would cause multiple floodings at the local level in the area.

\begin{figure*}[ht]
    \centering
    \includegraphics[width=0.75\textwidth]{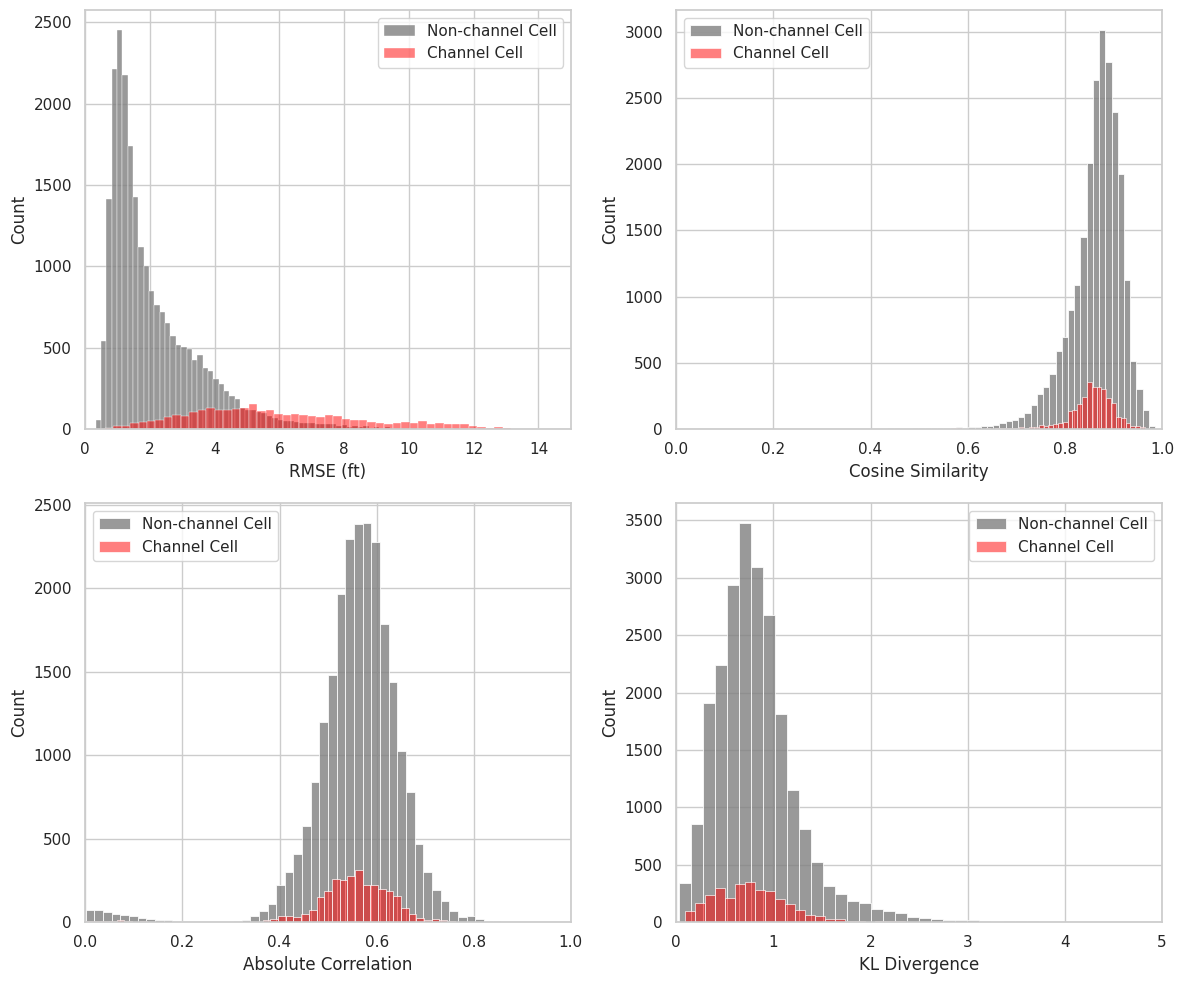}
    \caption{Histograms of Flood Assessment Metrics: Comparison of metrics between channel cells and non-channel cells, based on the differences between the sampled training depth distribution and the downsampled synthetic depth distribution.}
    \label{fig:r_vs_s}
\end{figure*}

\begin{table*}[]
    \centering
    \footnotesize
    \caption{Statistical comparison of sampled training depth distribution and downsampled synthetic depth distribution for key metrics across Overall, Channel, and Non-channel Cells.}
    \begin{tabular}{lcccc}
        \hline
        \textbf{Metric} & \textbf{RMSE (ft)} & \textbf{Cosine Similarity} & \textbf{Correlation} & \textbf{KL Divergence} \\
        \hline
        \textbf{Overall} \\
        Count & 26301 & 26301 & 26301 & 26301 \\
        Mean  & 2.5813 & 0.8537 & 0.5621 & 0.8332 \\
        Std   & 2.0925 & 0.0553 & 0.0996 & 0.4610 \\
        Min   & 0.3248 & 0.3523 & 0.0010 & 0.0236 \\
        25\%  & 1.1264 & 0.8452 & 0.5205 & 0.5308 \\
        50\%  & 1.8231 & 0.8627 & 0.5588 & 0.7692 \\
        75\%  & 3.3417 & 0.8911 & 0.6034 & 1.0227 \\
        Max   & 26.982 & 0.9923 & 0.8812 & 6.2701 \\
        \hline
        \textbf{Channel} \\
        Count & 2985 & 2985 & 2985 & 2985 \\
        Mean  & 5.9978 & 0.8461 & 0.5433 & 0.7831 \\
        Std   & 2.7955 & 0.0489 & 0.0998 & 0.4392 \\
        Min   & 0.6084 & 0.4735 & 0.0015 & 0.0905 \\
        25\%  & 3.9428 & 0.8276 & 0.5151 & 0.4697 \\
        50\%  & 5.4801 & 0.8662 & 0.5513 & 0.7598 \\
        75\%  & 7.7058 & 0.8785 & 0.6121 & 1.0152 \\
        Max   & 27.015 & 0.9578 & 0.8015 & 5.1892 \\
        \hline
        \textbf{Non-channel} \\
        Count & 23316 & 23316 & 23316 & 23316 \\
        Mean  & 2.0992 & 0.8696 & 0.5525 & 0.8224 \\
        Std   & 1.4678 & 0.0562 & 0.1013 & 0.4532 \\
        Min   & 0.3221 & 0.3511 & 0.0010 & 0.0240 \\
        25\%  & 1.0912 & 0.8415 & 0.5240 & 0.5521 \\
        50\%  & 1.6098 & 0.8772 & 0.5644 & 0.7559 \\
        75\%  & 2.7623 & 0.9028 & 0.6183 & 1.0194 \\
        Max   & 16.940 & 0.9815 & 0.8897 & 6.2550 \\
        \hline
    \end{tabular}
    
    \label{tab:true_vs_syn}
\end{table*}

\subsection{Synthetic Rainfall Distribution Assessment}
We generated 10,000 synthetic precipitation-flood events, each cell having a comprehensive synthetic depth distribution for direct comparison with the training dataset. To assess the performance of the synthetic events quantitatively, we utilized four key metrics—RMSE, Cosine Similarity, Correlation, and KL Divergence—comparing them against the limited number of sampled training events. To ensure a fair comparison, the synthetic events were downsampled to match the size of the training data. This downsampling process was repeated 50 times to minimize variability introduced by randomness. A detailed statistical comparison between the normalized versions of the training and the synthetic datasets is provided in Table \ref{tab:true_vs_syn}, with the corresponding visualizations shown in Fig. \ref{fig:r_vs_s}.

The sampled synthetic flood depth distribution demonstrates comparable overall performance, as evidenced by the RMSE and cosine similarity metrics, indicating that the generated events align closely with the training data and accurately capture key hydrological characteristics and trends. Notably, the mean RMSE for non-channel cells is 2.10 ft, substantially lower than the 5.99 ft recorded for channel cells. This disparity suggests that the synthetic data more effectively represents regions with simpler hydrodynamics, while the higher RMSE in channel cells reflects the inherent difficulty in modeling real-world depth distributions in areas characterized by dynamic flow patterns and greater variability. Nevertheless, the synthetic data performs well even in these more complex regions, achieving a mean cosine similarity of 0.85 for channel cells, highlighting the model's ability to capture the overall structure and trends of the real data.

In non-channel cells, the model excels with a mean cosine similarity of 0.87 and a relatively low KL divergence of 0.82. These metrics suggest that the synthetic data closely matches the probability distributions of the real data in these regions, where simpler hydrological behavior makes depth distributions easier to replicate. In contrast, the KL divergence for channel cells, though still low at 0.78, indicates some deviation between the synthetic and real data, which is expected given the complexity and variability of flood behavior in these more dynamic areas.

The correlation between the synthetic and real datasets further reinforces the model's effectiveness. Non-channel cells exhibit a mean correlation of 0.55, while channel cells show a slightly lower but still robust correlation of 0.54. This highlights the model’s ability to capture linear relationships across both cell types, though the more intricate hydrodynamic processes in channel cells present additional challenges.
Hydrologically, these findings are particularly significant. Channel cells, which are more prone to flooding due to their dynamic flow patterns, underscore the model's strength in capturing intricate interactions even under complex conditions. The model’s ability to generalize across non-channel cells, reflected in their lower RMSE and KL divergence values, is equally noteworthy. It demonstrates the model’s capacity to explore a variety of flood scenarios without overfitting to specific events.

Geographically, the clear distinction between channel and non-channel cells emphasizes the importance of local topographical and hydrological features in flood modeling. The synthetic data performs particularly well in regions with less complex hydrodynamics, while still maintaining a reasonable alignment in more intricate areas.

Overall, these results highlight the strong potential of the synthetic dataset for reliable flood depth estimation across diverse regions, especially in flood-prone channel areas where accurate modeling is crucial for effective risk assessment and mitigation. The close alignment between the synthetic and training data across multiple metrics underscores the robustness of the approach, demonstrating its ability to replicate complex flood patterns while maintaining generalizability across varied hydrological landscapes. These findings affirm the method’s suitability for predictive flood modeling and underscore its potential as a powerful tool for future flood risk assessment and management efforts.

\subsection{Flood probability maps}
\begin{figure*}[]
    \centering
    \begin{subfigure}[b]{0.48\textwidth}
        \includegraphics[width=\textwidth]{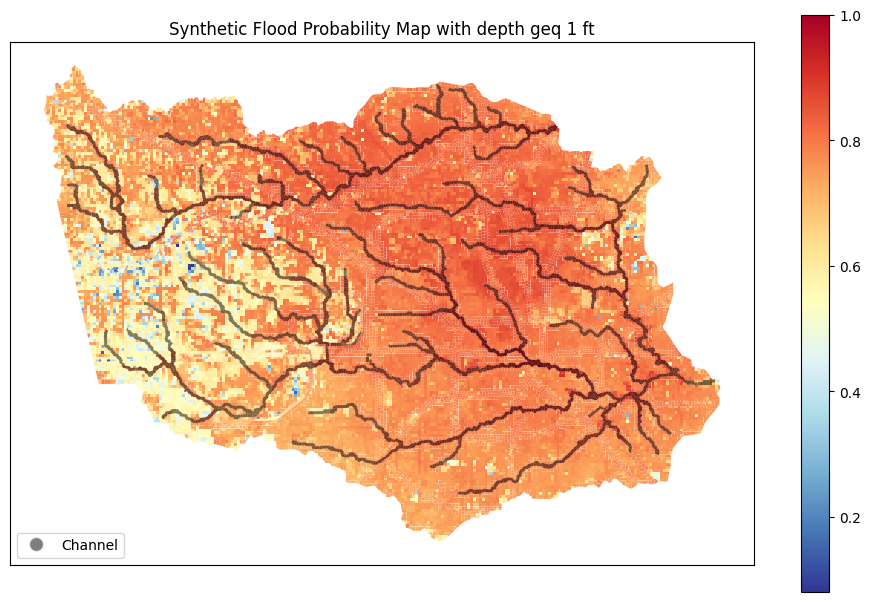}
        \caption{$P(\text{depth}\geq 1\text{ft})$}
    \end{subfigure}
    \begin{subfigure}[b]{0.48\textwidth}
        \includegraphics[width=\textwidth]{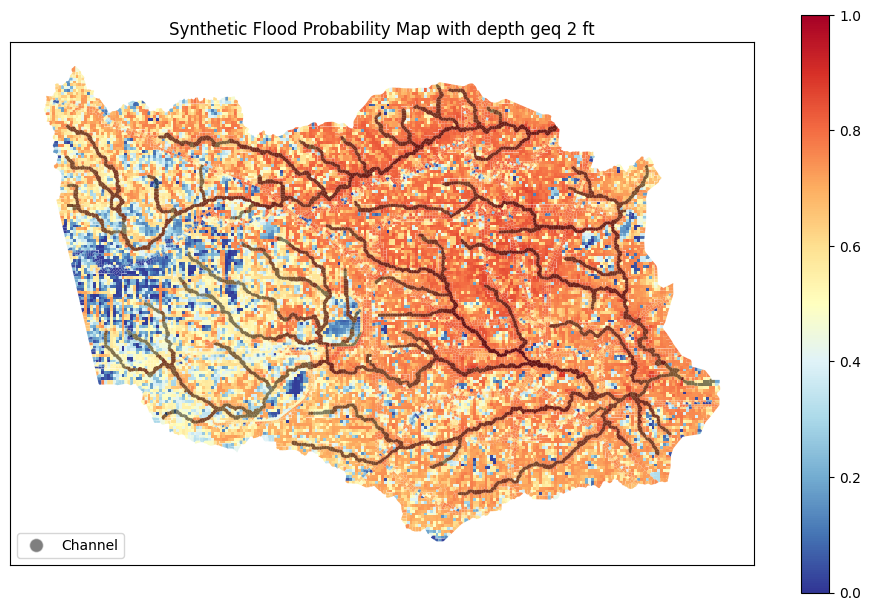}
        \caption{$P(\text{depth}\geq 2\text{ft})$}
    \end{subfigure}
    \newline
    \begin{subfigure}[b]{0.48\textwidth}
        \includegraphics[width=\textwidth]{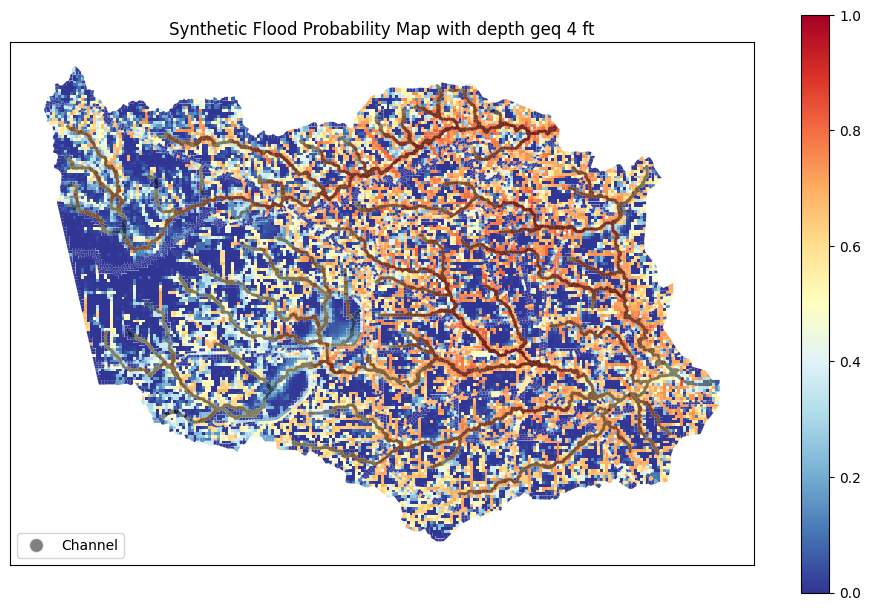}
        \caption{$P(\text{depth}\geq 4\text{ft})$}
    \end{subfigure}
    \begin{subfigure}[b]{0.48\textwidth}
        \includegraphics[width=\textwidth]{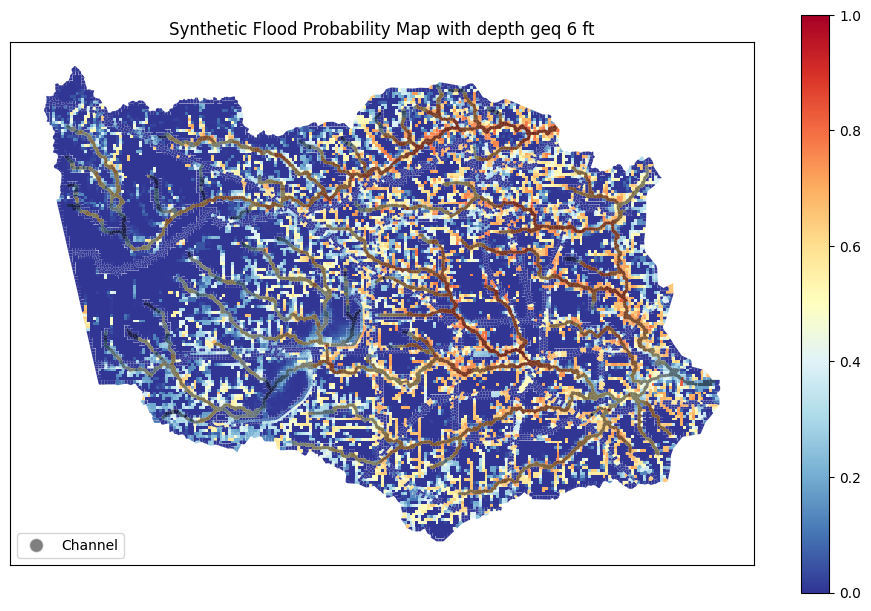}
        \caption{$P(\text{depth}\geq 6\text{ft})$}
    \end{subfigure}
    \caption{Synthetic flood probability maps based on four inundation depth criteria. The color gradient from blue to red represents probabilities ranging from 0 to 1 for a global precipitation event, with blue indicating lower probabilities and red indicating higher probabilities. Channel cells are highlighted in gray. Maps generated by \textit{Geopandas} Python package.}
    \label{fig:prob_maps}
\end{figure*}

Flood depths generated by a large number of synthetic precipitation-flood events represent a foundational data set from which one can estimate flood probability. Probability is calculated cell-wise as the proportion of synthetic events that cause a certain value of flood depth or higher. Using the 10,000 synthetic events simulated in Section \ref{sec:synth_rainfall}, we built synthetic flood probability maps for the following inundation depths: 1 ft, 2 ft, 4 ft, and 6 ft (Fig. \ref{fig:prob_maps}). In Fig. 6.IV, for example, a high $P(\text{depth}\geq 6 ft)$ in a cell means that for a generic flood event there is a high probability that flood level will equal or exceed 6 ft in that cell. In other words, the probability value serves as a quantification of risk for a certain flooding depth. This interpretation should be distinct from annual flood probability which is traditionally used in flood mapping. As shown in Fig. \ref{fig:prob_maps}, at 1 ft and 2 ft depths, flood probabilities greater than 0.5 (high probability) are predominantly observed in the northeastern, southeastern, and central regions of the study area, while lower probabilities are seen in the western and southern areas. As the depth threshold increases to 4 ft and 6 ft, the high-probability zones become more concentrated along rivers/channels and low-lying regions.

%% file: discussion.tex
\section{DISCUSSION}\label{sec:4}
In this study, we presented an innovative application methodology, Precipitation-Flood Depth Generative Pipeline, for generating high-resolution flood probability maps using synthetic precipitation-flood events. Our approach leveraged advanced machine learning techniques, including Conditional Generative Adversarial Networks (CTGAN) \citep{ctgan} as point location, and precipitation-based features generator and XGBoost-based models MaxFloodCast \citep{lee2024predicting} as synthetic depth estimator, to overcome the limitations of traditional flood modeling that rely on historical data and records, not always available. To address the limitations in feature dimensionality found in existing CTGAN tools and the challenge of generating individual precipitation-flood events from a massive synthetic point cloud, we designed a robust and systematic workflow. This includes cell-level pool creation, strategic sampling, noise smoothing, depth synthesis, and the eventual formation of individual events. Each step in the process is carefully structured to avoid overfitting and ensure that the generated synthetic events accurately reflect real-world flood scenarios while maintaining computational efficiency and high fidelity to hydrological patterns. By generating 10,000 synthetic precipitation-flood events, we created a comprehensive dataset that captures a wide spectrum of potential flood scenarios, increasing the robustness of flood risk assessments. The RMSE values highlight a notable difference between channel and non-channel cells, reflecting the challenges in accurately modeling flood depth distribution in regions with dynamic flows and greater variability. These findings demonstrate the model's ability to capture key trends and non-linear relationships, particularly along channels and in floodplain areas, which are more susceptible to flooding. Additionally, the synthetic data’s capacity to generalize beyond the training events highlights its potential for producing possible flood scenarios, which is critical for accurate risk assessment and mitigation planning.

\subsection{Spatial nature of weather and flooding}
Despite the advancements introduced with the Precipitation-Flood Depth Generative Pipeline, this approach has some limitations in generating local precipitation events. While CTGAN can generate records with no precipitation, we observed a tendency to generate storms that affect the majority of the study area. The prevalence of large-extent storms in the synthetic dataset influenced flood depth distributions at the cell level and consequently the flood probability maps. Flood probability maps currently in use in the study area show the extent of the 100-year and 500-year floodplain, but do not provide information on associated flood depths at the largest scale, making difficult a direct comparison with the synthetic flood probability maps illustrated in Fig. \ref{fig:prob_maps}. However, the flood probability for low depth values (1 and 2 feet) is likely overestimated outside of channels, given the spatial uniformity of high values, even on elevated road cells and watershed divides. For the higher depths considered (4 and 6 feet), synthetic flood probabilities have more distinct values between low-lying and high-lying areas, and align more closely to the latest 100-year and 500-year floodplain extents \citep{web_floodquilt}.  

Synthetic flood probability overestimation can be explained because the three storm characteristics, i.e., duration, cumulative rainfall and peak intensity, are generated at the cell level independently, without considering the spatial autocorrelation existing in an actual storm. Future work will focus on additional characteristics that capture the spatial nature of precipitation events. These include the spatial autocorrelation of storm events of different duration, intensity and extent, the directions along which storms typically moves, and other spatio-temporal properties defined at the event-scale rather than cell-wise, e.g., rotation and velocity.

\subsection{Computational efficiency}
The computational complexity of the pipeline is overall significant. The cell-wise depth estimator costs higher computational resource than the universal models tested (Section \ref{sec:estimator_select}), taking approximately 10 minutes and 100GB CPU memory to generate 100 precipitation-flood events. This performance is however competitive over HEC-RAS 2D, for which \cite{lee2024predicting} found that the runtime for a single simulation was around 3 minutes, depending on the storm extent and duration. The obvious advantage of MaxFloodCast is the cell-model parallelization, which is not possible in HEC-RAS 2D. Future work should improve the efficiency of the computational process, especially in presence of additional relationships of spatial and hydrological nature that would enhance precipitation event generation.

\subsection{Applications and future research}
The Precipitation-Flood Depth Generative Pipeline was applied in Harris County due to the availability of an initial dataset of storms for training, as well as of a calibrated physics-based model that provided the data used to train a depth-estimator. The availability of historical storms and the resulting flood depth levels is necessary for the pipeline application in any study area, whether it is a prevalently natural, rural or urbanized region. Different model parametrization might be necessary to adapt to region-specific meteorological and hydrological conditions, especially when generating synthetic storms. Future research should determine what the appropriate scale is for the pipeline application, and if the scale difference between weather analysis (regional scale) and flood depth analysis (watershed scale) represents an issue for implementing the pipeline and producing accurate flood probability maps. 

Flood probability maps provide a spatial analysis of flood risk as a function of water depth. These maps emphasize the importance of depth-specific flood management strategies. This is crucial in regions where intricate flow dynamics and a flat ground surface significantly influence flood behavior. The high resolution of the maps also allows for practical applications, such as assessing the flood risk of buildings by integrating lowest floor elevation datasets, or identifying structures at risk of inundation a certain threshold. Additionally, these maps can evaluate transportation network vulnerability, as roads in areas with a high probability of exceeding 2 ft of inundation would become impassable, disrupting transportation and emergency response efforts. The non-binary nature of the maps further allows for comparative flood risk assessments across different inundation levels, enabling more nuanced and targeted flood mitigation strategies across diverse landscapes.

Finally, this study also contributes to efforts to expand synthetic data for AI models developed for disaster resilience use cases. Both real and synthetic data support the development of robust AI models for various disaster resilience use cases ranging from risk prediction to situational awareness and impact assessment. Future studies can leverage GAN and other generative ML methods for augmenting data from occurred disasters and produce datasets suitable for training and testing novel ML models in multiple disaster resilience applications.

%% file: conclusion.tex
\section{Conclusions}
In this study, we proposed the Precipitation-Flood Depth Generative Pipeline, a novel machine learning framework for generating synthetic precipitation-flood events to construct high-resolution probabilistic flood maps. By integrating a Conditional Generative Adversarial Network (CTGAN) for synthetic precipitation generation with a specialized cell-wise depth estimator (MaxFloodCast V2), our approach addresses some of the limitations in traditional flood mapping methods, including data scarcity and computational costs associated with physics-based simulations. The results show the capability of the pipeline to produce flood probability maps that align with hydrological intuition and known flood-prone areas, highlighting its potential as a scalable tool for flood risk assessment.

Despite its advantages, certain limitations remain, such as potential inconsistencies in extreme precipitation cases and challenges in fully capturing spatial dependencies. Future research will focus on enhancing the integration of generative models with physics-based simulations, improving event-generation mechanisms to better capture localized precipitations, and further benchmarking against alternative deep learning models. Additionally, extending this framework to different geographical regions and incorporating real-time flood forecasting applications could expand its utility in disaster resilience and urban planning.

%% file: wileyNJD-APA.bbl
\begin{thebibliography}{}

\bibitem [\protect \citeauthoryear {%
Adams%
, Southard%
, Forbes%
, Leinweber%
\BCBL {}\ \BBA {} Nix%
}{%
Adams%
\ \protect \BOthers {.}}{%
{\protect \APACyear {2022}}%
}]{%
adams2022mitigation}
\APACinsertmetastar {%
adams2022mitigation}%
\begin{APACrefauthors}%
Adams, H.%
, Southard, M.%
, Forbes, S.%
, Leinweber, B.%
\BCBL {}\ \BBA {} Nix, D.%
\end{APACrefauthors}%
\unskip\
\newblock
\APACrefYearMonthDay{2022}{01}{}.
\newblock
{\BBOQ}\APACrefatitle {Effective Training Improves Disaster Response} {Effective training improves disaster response}.{\BBCQ}
\newblock
\APACjournalVolNumPages{Opflow}{48}{}{10-14}.
\newblock
\begin{APACrefDOI} \doi{10.1002/opfl.1632} \end{APACrefDOI}
\PrintBackRefs{\CurrentBib}

\bibitem [\protect \citeauthoryear {%
Castangia%
\ \protect \BOthers {.}}{%
Castangia%
\ \protect \BOthers {.}}{%
{\protect \APACyear {2023}}%
}]{%
castangia2023transformer}
\APACinsertmetastar {%
castangia2023transformer}%
\begin{APACrefauthors}%
Castangia, M.%
, Grajales, L\BPBI M\BPBI M.%
, Aliberti, A.%
, Rossi, C.%
, Macii, A.%
, Macii, E.%
\BCBL {}\ \BBA {} Patti, E.%
\end{APACrefauthors}%
\unskip\
\newblock
\APACrefYearMonthDay{2023}{}{}.
\newblock
{\BBOQ}\APACrefatitle {Transformer neural networks for interpretable flood forecasting} {Transformer neural networks for interpretable flood forecasting}.{\BBCQ}
\newblock
\APACjournalVolNumPages{Environmental Modelling \& Software}{160}{}{105581}.
\PrintBackRefs{\CurrentBib}

\bibitem [\protect \citeauthoryear {%
Chen%
\ \BBA {} Guestrin%
}{%
Chen%
\ \BBA {} Guestrin%
}{%
{\protect \APACyear {2016}}%
}]{%
chen2016xgboost}
\APACinsertmetastar {%
chen2016xgboost}%
\begin{APACrefauthors}%
Chen, T.%
\BCBT {}\ \BBA {} Guestrin, C.%
\end{APACrefauthors}%
\unskip\
\newblock
\APACrefYearMonthDay{2016}{}{}.
\newblock
{\BBOQ}\APACrefatitle {Xgboost: A scalable tree boosting system} {Xgboost: A scalable tree boosting system}.{\BBCQ}
\newblock
\BIn{} \APACrefbtitle {Proceedings of the 22nd acm sigkdd international conference on knowledge discovery and data mining} {Proceedings of the 22nd acm sigkdd international conference on knowledge discovery and data mining}\ (\BPGS\ 785--794).
\PrintBackRefs{\CurrentBib}

\bibitem [\protect \citeauthoryear {%
Chicco%
, Warrens%
\BCBL {}\ \BBA {} Jurman%
}{%
Chicco%
\ \protect \BOthers {.}}{%
{\protect \APACyear {2021}}%
}]{%
chicco2021coefficient}
\APACinsertmetastar {%
chicco2021coefficient}%
\begin{APACrefauthors}%
Chicco, D.%
, Warrens, M\BPBI J.%
\BCBL {}\ \BBA {} Jurman, G.%
\end{APACrefauthors}%
\unskip\
\newblock
\APACrefYearMonthDay{2021}{}{}.
\newblock
{\BBOQ}\APACrefatitle {The coefficient of determination R-squared is more informative than SMAPE, MAE, MAPE, MSE and RMSE in regression analysis evaluation} {The coefficient of determination r-squared is more informative than smape, mae, mape, mse and rmse in regression analysis evaluation}.{\BBCQ}
\newblock
\APACjournalVolNumPages{Peerj computer science}{7}{}{e623}.
\PrintBackRefs{\CurrentBib}

\bibitem [\protect \citeauthoryear {%
Dewitz%
}{%
Dewitz%
}{%
{\protect \APACyear {2021}}%
}]{%
dewitzNationalLandCover2021}
\APACinsertmetastar {%
dewitzNationalLandCover2021}%
\begin{APACrefauthors}%
Dewitz, J.%
\end{APACrefauthors}%
\unskip\
\newblock
\APACrefYearMonthDay{2021}{}{}.
\newblock
\APACrefbtitle {National {{Land Cover Database}} ({{NLCD}}) 2019 {{Products}}.} {National {{Land Cover Database}} ({{NLCD}}) 2019 {{Products}}.}
\newblock
\APACaddressPublisher{}{{U.S. Geological Survey}}.
\newblock
\begin{APACrefDOI} \doi{10.5066/P9KZCM54} \end{APACrefDOI}
\PrintBackRefs{\CurrentBib}

\bibitem [\protect \citeauthoryear {%
{do Lago}%
\ \protect \BOthers {.}}{%
{do Lago}%
\ \protect \BOthers {.}}{%
{\protect \APACyear {2023}}%
}]{%
DOLAGO2023129276}
\APACinsertmetastar {%
DOLAGO2023129276}%
\begin{APACrefauthors}%
{do Lago}, C\BPBI A.%
, Giacomoni, M\BPBI H.%
, Bentivoglio, R.%
, Taormina, R.%
, Gomes, M\BPBI N.%
\BCBL {}\ \BBA {} Mendiondo, E\BPBI M.%
\end{APACrefauthors}%
\unskip\
\newblock
\APACrefYearMonthDay{2023}{}{}.
\newblock
{\BBOQ}\APACrefatitle {Generalizing rapid flood predictions to unseen urban catchments with conditional generative adversarial networks} {Generalizing rapid flood predictions to unseen urban catchments with conditional generative adversarial networks}.{\BBCQ}
\newblock
\APACjournalVolNumPages{Journal of Hydrology}{618}{}{129276}.
\newblock
\begin{APACrefURL} \url{https://www.sciencedirect.com/science/article/pii/S0022169423002184} \end{APACrefURL}
\newblock
\begin{APACrefDOI} \doi{https://doi.org/10.1016/j.jhydrol.2023.129276} \end{APACrefDOI}
\PrintBackRefs{\CurrentBib}

\bibitem [\protect \citeauthoryear {%
Freedman%
, Pisani%
\BCBL {}\ \BBA {} Purves%
}{%
Freedman%
\ \protect \BOthers {.}}{%
{\protect \APACyear {2007}}%
}]{%
freedman2007statistics}
\APACinsertmetastar {%
freedman2007statistics}%
\begin{APACrefauthors}%
Freedman, D.%
, Pisani, R.%
\BCBL {}\ \BBA {} Purves, R.%
\end{APACrefauthors}%
\unskip\
\newblock
\APACrefYearMonthDay{2007}{}{}.
\newblock
{\BBOQ}\APACrefatitle {Statistics (international student edition)} {Statistics (international student edition)}.{\BBCQ}
\newblock
\APACjournalVolNumPages{Pisani, R. Purves, 4th edn. WW Norton \& Company, New York}{}{}{}.
\PrintBackRefs{\CurrentBib}

\bibitem [\protect \citeauthoryear {%
M.~Garcia%
, Juan%
, {Doss-Gollin}%
\BCBL {}\ \BBA {} Bedient%
}{%
M.~Garcia%
\ \protect \BOthers {.}}{%
{\protect \APACyear {2023}}%
}]{%
garciaLeveragingMeshModularization2023}
\APACinsertmetastar {%
garciaLeveragingMeshModularization2023}%
\begin{APACrefauthors}%
Garcia, M.%
, Juan, A.%
, {Doss-Gollin}, J.%
\BCBL {}\ \BBA {} Bedient, P.%
\end{APACrefauthors}%
\unskip\
\newblock
\APACrefYearMonthDay{2023}{{\APACmonth{12}}}{}.
\newblock
{\BBOQ}\APACrefatitle {Leveraging Mesh Modularization to Lower the Computational Cost of Localized Updates to Regional {{2D}} Hydrodynamic Model Outputs} {Leveraging mesh modularization to lower the computational cost of localized updates to regional {{2D}} hydrodynamic model outputs}.{\BBCQ}
\newblock
\APACjournalVolNumPages{Engineering Applications of Computational Fluid Mechanics}{17}{1}{2225584}.
\newblock
\begin{APACrefDOI} \doi{10.1080/19942060.2023.2225584} \end{APACrefDOI}
\PrintBackRefs{\CurrentBib}

\bibitem [\protect \citeauthoryear {%
M\BPBI S.~Garcia%
}{%
M\BPBI S.~Garcia%
}{%
{\protect \APACyear {2023}}%
}]{%
garciaphdthesis2023}
\APACinsertmetastar {%
garciaphdthesis2023}%
\begin{APACrefauthors}%
Garcia, M\BPBI S.%
\end{APACrefauthors}%
\unskip\
\newblock
\APACrefYear{2023}.
\unskip\
\newblock
\APACrefbtitle {Novel Urban Floodplain Modeling Methods for Applications in Coupling Surrogate Machine Learning Methods} {Novel urban floodplain modeling methods for applications in coupling surrogate machine learning methods}\ \APACtypeAddressSchool {Doctoral Dissertation}{}{Rice University}.
\unskip\
\newblock
\begin{APACrefURL} \url{https://hdl.handle.net/1911/115079} \end{APACrefURL}
\PrintBackRefs{\CurrentBib}

\bibitem [\protect \citeauthoryear {%
{'Harris County Flood Control District'}%
}{%
{'Harris County Flood Control District'}%
}{%
{\protect \APACyear {2023}}%
}]{%
harriscountyfloodcontroldistrictHarrisCountyFlood2023}
\APACinsertmetastar {%
harriscountyfloodcontroldistrictHarrisCountyFlood2023}%
\begin{APACrefauthors}%
{'Harris County Flood Control District'}.%
\end{APACrefauthors}%
\unskip\
\newblock
\APACrefYearMonthDay{2023}{}{}.
\newblock
\APACrefbtitle {Harris {{County Flood Warning System}}.} {Harris {{County Flood Warning System}}.}
\newblock
\begin{APACrefURL} \url{https://www.harriscountyfws.org/} \end{APACrefURL}
\PrintBackRefs{\CurrentBib}

\bibitem [\protect \citeauthoryear {%
Hershey%
\ \BBA {} Olsen%
}{%
Hershey%
\ \BBA {} Olsen%
}{%
{\protect \APACyear {2007}}%
}]{%
hershey2007approximating}
\APACinsertmetastar {%
hershey2007approximating}%
\begin{APACrefauthors}%
Hershey, J\BPBI R.%
\BCBT {}\ \BBA {} Olsen, P\BPBI A.%
\end{APACrefauthors}%
\unskip\
\newblock
\APACrefYearMonthDay{2007}{}{}.
\newblock
{\BBOQ}\APACrefatitle {Approximating the Kullback Leibler divergence between Gaussian mixture models} {Approximating the kullback leibler divergence between gaussian mixture models}.{\BBCQ}
\newblock
\BIn{} \APACrefbtitle {2007 IEEE International Conference on Acoustics, Speech and Signal Processing-ICASSP'07} {2007 ieee international conference on acoustics, speech and signal processing-icassp'07}\ (\BVOL~4, \BPGS\ IV--317).
\PrintBackRefs{\CurrentBib}

\bibitem [\protect \citeauthoryear {%
Ho%
, Li%
\BCBL {}\ \BBA {} Mostafavi%
}{%
Ho%
\ \protect \BOthers {.}}{%
{\protect \APACyear {{\protect \bibnodate {}}}}%
}]{%
ho2024integrated}
\APACinsertmetastar {%
ho2024integrated}%
\begin{APACrefauthors}%
Ho, Y\BHBI H.%
, Li, L.%
\BCBL {}\ \BBA {} Mostafavi, A.%
\end{APACrefauthors}%
\unskip\
\newblock
\APACrefYearMonthDay{{\protect \bibnodate {}}}{}{}.
\newblock
{\BBOQ}\APACrefatitle {Integrated vision language and foundation model for automated estimation of building lowest floor elevation} {Integrated vision language and foundation model for automated estimation of building lowest floor elevation}.{\BBCQ}
\newblock
\APACjournalVolNumPages{Computer-Aided Civil and Infrastructure Engineering}{}{}{}.
\PrintBackRefs{\CurrentBib}

\bibitem [\protect \citeauthoryear {%
Ji%
, Mirzaei%
, Lai%
, Dehghani%
\BCBL {}\ \BBA {} Dehghani%
}{%
Ji%
\ \protect \BOthers {.}}{%
{\protect \APACyear {2024}}%
}]{%
JI2024105896}
\APACinsertmetastar {%
JI2024105896}%
\begin{APACrefauthors}%
Ji, H\BPBI K.%
, Mirzaei, M.%
, Lai, S\BPBI H.%
, Dehghani, A.%
\BCBL {}\ \BBA {} Dehghani, A.%
\end{APACrefauthors}%
\unskip\
\newblock
\APACrefYearMonthDay{2024}{}{}.
\newblock
{\BBOQ}\APACrefatitle {Implementing generative adversarial network (GAN) as a data-driven multi-site stochastic weather generator for flood frequency estimation} {Implementing generative adversarial network (gan) as a data-driven multi-site stochastic weather generator for flood frequency estimation}.{\BBCQ}
\newblock
\APACjournalVolNumPages{Environmental Modelling \& Software}{172}{}{105896}.
\newblock
\begin{APACrefURL} \url{https://www.sciencedirect.com/science/article/pii/S1364815223002827} \end{APACrefURL}
\newblock
\begin{APACrefDOI} \doi{https://doi.org/10.1016/j.envsoft.2023.105896} \end{APACrefDOI}
\PrintBackRefs{\CurrentBib}

\bibitem [\protect \citeauthoryear {%
Konrad%
}{%
Konrad%
}{%
{\protect \APACyear {2021}}%
}]{%
markus_konrad_2021_4531339}
\APACinsertmetastar {%
markus_konrad_2021_4531339}%
\begin{APACrefauthors}%
Konrad, M.%
\end{APACrefauthors}%
\unskip\
\newblock
\APACrefYearMonthDay{2021}{{\APACmonth{02}}}{}.
\newblock
\APACrefbtitle {geovoronoi.} {geovoronoi.}
\newblock
\APACaddressPublisher{}{Zenodo}.
\newblock
\begin{APACrefURL} \url{https://doi.org/10.5281/zenodo.4531339} \end{APACrefURL}
\newblock
\begin{APACrefDOI} \doi{10.5281/zenodo.4531339} \end{APACrefDOI}
\PrintBackRefs{\CurrentBib}

\bibitem [\protect \citeauthoryear {%
Lee%
\ \protect \BOthers {.}}{%
Lee%
\ \protect \BOthers {.}}{%
{\protect \APACyear {2024}}%
}]{%
lee2024predicting}
\APACinsertmetastar {%
lee2024predicting}%
\begin{APACrefauthors}%
Lee, C\BHBI C.%
, Huang, L.%
, Antolini, F.%
, Garcia, M.%
, Juan, A.%
, Brody, S\BPBI D.%
\BCBL {}\ \BBA {} Mostafavi, A.%
\end{APACrefauthors}%
\unskip\
\newblock
\APACrefYearMonthDay{2024}{}{}.
\newblock
{\BBOQ}\APACrefatitle {Predicting peak inundation depths with a physics informed machine learning model} {Predicting peak inundation depths with a physics informed machine learning model}.{\BBCQ}
\newblock
\APACjournalVolNumPages{Scientific Reports}{14}{1}{14826}.
\PrintBackRefs{\CurrentBib}

\bibitem [\protect \citeauthoryear {%
Lyu%
\ \BBA {} Yin%
}{%
Lyu%
\ \BBA {} Yin%
}{%
{\protect \APACyear {2023}}%
}]{%
LYU2023104744}
\APACinsertmetastar {%
LYU2023104744}%
\begin{APACrefauthors}%
Lyu, H\BHBI M.%
\BCBT {}\ \BBA {} Yin, Z\BHBI Y.%
\end{APACrefauthors}%
\unskip\
\newblock
\APACrefYearMonthDay{2023}{}{}.
\newblock
{\BBOQ}\APACrefatitle {Flood susceptibility prediction using tree-based machine learning models in the GBA} {Flood susceptibility prediction using tree-based machine learning models in the gba}.{\BBCQ}
\newblock
\APACjournalVolNumPages{Sustainable Cities and Society}{97}{}{104744}.
\newblock
\begin{APACrefURL} \url{https://www.sciencedirect.com/science/article/pii/S2210670723003554} \end{APACrefURL}
\newblock
\begin{APACrefDOI} \doi{https://doi.org/10.1016/j.scs.2023.104744} \end{APACrefDOI}
\PrintBackRefs{\CurrentBib}

\bibitem [\protect \citeauthoryear {%
McSpadden%
\ \protect \BOthers {.}}{%
McSpadden%
\ \protect \BOthers {.}}{%
{\protect \APACyear {2024}}%
}]{%
MCSPADDEN2024100518}
\APACinsertmetastar {%
MCSPADDEN2024100518}%
\begin{APACrefauthors}%
McSpadden, D.%
, Goldenberg, S.%
, Roy, B.%
, Schram, M.%
, Goodall, J\BPBI L.%
\BCBL {}\ \BBA {} Richter, H.%
\end{APACrefauthors}%
\unskip\
\newblock
\APACrefYearMonthDay{2024}{}{}.
\newblock
{\BBOQ}\APACrefatitle {A comparison of machine learning surrogate models of street-scale flooding in Norfolk, Virginia} {A comparison of machine learning surrogate models of street-scale flooding in norfolk, virginia}.{\BBCQ}
\newblock
\APACjournalVolNumPages{Machine Learning with Applications}{15}{}{100518}.
\newblock
\begin{APACrefURL} \url{https://www.sciencedirect.com/science/article/pii/S2666827023000713} \end{APACrefURL}
\newblock
\begin{APACrefDOI} \doi{https://doi.org/10.1016/j.mlwa.2023.100518} \end{APACrefDOI}
\PrintBackRefs{\CurrentBib}

\bibitem [\protect \citeauthoryear {%
Monge%
}{%
Monge%
}{%
{\protect \APACyear {2023}}%
}]{%
monge2023two}
\APACinsertmetastar {%
monge2023two}%
\begin{APACrefauthors}%
Monge, M.%
\end{APACrefauthors}%
\unskip\
\newblock
\APACrefYearMonthDay{2023}{}{}.
\newblock
{\BBOQ}\APACrefatitle {Two-Sample Kolmogorov-Smirnov Tests as Causality Tests. A Narrative of Latin American inflation from 2020 to 2022.} {Two-sample kolmogorov-smirnov tests as causality tests. a narrative of latin american inflation from 2020 to 2022.}{\BBCQ}
\newblock
\APACjournalVolNumPages{Revista Chilena de Econom{\'\i}a y Sociedad}{}{}{}.
\PrintBackRefs{\CurrentBib}

\bibitem [\protect \citeauthoryear {%
Musselman%
\ \protect \BOthers {.}}{%
Musselman%
\ \protect \BOthers {.}}{%
{\protect \APACyear {2018}}%
}]{%
musselman2018projected}
\APACinsertmetastar {%
musselman2018projected}%
\begin{APACrefauthors}%
Musselman, K\BPBI N.%
, Lehner, F.%
, Ikeda, K.%
, Clark, M\BPBI P.%
, Prein, A\BPBI F.%
, Liu, C.%
\BDBL {}Rasmussen, R.%
\end{APACrefauthors}%
\unskip\
\newblock
\APACrefYearMonthDay{2018}{}{}.
\newblock
{\BBOQ}\APACrefatitle {Projected increases and shifts in rain-on-snow flood risk over western North America} {Projected increases and shifts in rain-on-snow flood risk over western north america}.{\BBCQ}
\newblock
\APACjournalVolNumPages{Nature Climate Change}{8}{9}{808--812}.
\PrintBackRefs{\CurrentBib}

\bibitem [\protect \citeauthoryear {%
Nofal%
\ \protect \BOthers {.}}{%
Nofal%
\ \protect \BOthers {.}}{%
{\protect \APACyear {2024}}%
}]{%
nofal2024community}
\APACinsertmetastar {%
nofal2024community}%
\begin{APACrefauthors}%
Nofal, O.%
, Rosenheim, N.%
, Kameshwar, S.%
, Patil, J.%
, Zhou, X.%
, van~de Lindt, J\BPBI W.%
\BDBL {}others%
\end{APACrefauthors}%
\unskip\
\newblock
\APACrefYearMonthDay{2024}{}{}.
\newblock
{\BBOQ}\APACrefatitle {Community-level post-hazard functionality methodology for buildings exposed to floods} {Community-level post-hazard functionality methodology for buildings exposed to floods}.{\BBCQ}
\newblock
\APACjournalVolNumPages{Computer-Aided Civil and Infrastructure Engineering}{39}{8}{1099--1122}.
\PrintBackRefs{\CurrentBib}

\bibitem [\protect \citeauthoryear {%
Patki%
, Wedge%
\BCBL {}\ \BBA {} Veeramachaneni%
}{%
Patki%
\ \protect \BOthers {.}}{%
{\protect \APACyear {2016}}%
}]{%
SDV}
\APACinsertmetastar {%
SDV}%
\begin{APACrefauthors}%
Patki, N.%
, Wedge, R.%
\BCBL {}\ \BBA {} Veeramachaneni, K.%
\end{APACrefauthors}%
\unskip\
\newblock
\APACrefYearMonthDay{2016}{Oct}{}.
\newblock
{\BBOQ}\APACrefatitle {The Synthetic data vault} {The synthetic data vault}.{\BBCQ}
\newblock
\BIn{} \APACrefbtitle {IEEE International Conference on Data Science and Advanced Analytics (DSAA)} {Ieee international conference on data science and advanced analytics (dsaa)}\ (\BPG~399-410).
\newblock
\begin{APACrefDOI} \doi{10.1109/DSAA.2016.49} \end{APACrefDOI}
\PrintBackRefs{\CurrentBib}

\bibitem [\protect \citeauthoryear {%
Perica%
\ \protect \BOthers {.}}{%
Perica%
\ \protect \BOthers {.}}{%
{\protect \APACyear {2018}}%
}]{%
perica2018precipitation}
\APACinsertmetastar {%
perica2018precipitation}%
\begin{APACrefauthors}%
Perica, S.%
, Pavlovic, S.%
, St~Laurent, M.%
, Trypaluk, C.%
, Unruh, D.%
\BCBL {}\ \BBA {} Wilhite, O.%
\end{APACrefauthors}%
\unskip\
\newblock
\APACrefYearMonthDay{2018}{}{}.
\newblock
{\BBOQ}\APACrefatitle {NOAA Atlas 14 Volume 11 Version 2} {Noaa atlas 14 volume 11 version 2}.{\BBCQ}
\newblock
\APACjournalVolNumPages{Precipitation-Frequency Atlas of the United States, Texas}{}{}{}.
\PrintBackRefs{\CurrentBib}

\bibitem [\protect \citeauthoryear {%
Piadeh%
\ \protect \BOthers {.}}{%
Piadeh%
\ \protect \BOthers {.}}{%
{\protect \APACyear {2023}}%
}]{%
PIADEH2023105772}
\APACinsertmetastar {%
PIADEH2023105772}%
\begin{APACrefauthors}%
Piadeh, F.%
, Behzadian, K.%
, Chen, A\BPBI S.%
, Campos, L\BPBI C.%
, Rizzuto, J\BPBI P.%
\BCBL {}\ \BBA {} Kapelan, Z.%
\end{APACrefauthors}%
\unskip\
\newblock
\APACrefYearMonthDay{2023}{}{}.
\newblock
{\BBOQ}\APACrefatitle {Event-based decision support algorithm for real-time flood forecasting in urban drainage systems using machine learning modelling} {Event-based decision support algorithm for real-time flood forecasting in urban drainage systems using machine learning modelling}.{\BBCQ}
\newblock
\APACjournalVolNumPages{Environmental Modelling \& Software}{167}{}{105772}.
\newblock
\begin{APACrefURL} \url{https://www.sciencedirect.com/science/article/pii/S1364815223001585} \end{APACrefURL}
\newblock
\begin{APACrefDOI} \doi{https://doi.org/10.1016/j.envsoft.2023.105772} \end{APACrefDOI}
\PrintBackRefs{\CurrentBib}

\bibitem [\protect \citeauthoryear {%
{Saviz Naeini}%
\ \BBA {} Snaiki%
}{%
{Saviz Naeini}%
\ \BBA {} Snaiki%
}{%
{\protect \APACyear {2024}}%
}]{%
SAVIZNAEINI2024116986}
\APACinsertmetastar {%
SAVIZNAEINI2024116986}%
\begin{APACrefauthors}%
{Saviz Naeini}, S.%
\BCBT {}\ \BBA {} Snaiki, R.%
\end{APACrefauthors}%
\unskip\
\newblock
\APACrefYearMonthDay{2024}{}{}.
\newblock
{\BBOQ}\APACrefatitle {A physics-informed machine learning model for time-dependent wave runup prediction} {A physics-informed machine learning model for time-dependent wave runup prediction}.{\BBCQ}
\newblock
\APACjournalVolNumPages{Ocean Engineering}{295}{}{116986}.
\newblock
\begin{APACrefURL} \url{https://www.sciencedirect.com/science/article/pii/S0029801824003238} \end{APACrefURL}
\newblock
\begin{APACrefDOI} \doi{https://doi.org/10.1016/j.oceaneng.2024.116986} \end{APACrefDOI}
\PrintBackRefs{\CurrentBib}

\bibitem [\protect \citeauthoryear {%
Shehadeh%
, Alshboul%
, Al~Mamlook%
\BCBL {}\ \BBA {} Hamedat%
}{%
Shehadeh%
\ \protect \BOthers {.}}{%
{\protect \APACyear {2021}}%
}]{%
shehadeh2021machine}
\APACinsertmetastar {%
shehadeh2021machine}%
\begin{APACrefauthors}%
Shehadeh, A.%
, Alshboul, O.%
, Al~Mamlook, R\BPBI E.%
\BCBL {}\ \BBA {} Hamedat, O.%
\end{APACrefauthors}%
\unskip\
\newblock
\APACrefYearMonthDay{2021}{}{}.
\newblock
{\BBOQ}\APACrefatitle {Machine learning models for predicting the residual value of heavy construction equipment: An evaluation of modified decision tree, LightGBM, and XGBoost regression} {Machine learning models for predicting the residual value of heavy construction equipment: An evaluation of modified decision tree, lightgbm, and xgboost regression}.{\BBCQ}
\newblock
\APACjournalVolNumPages{Automation in Construction}{129}{}{103827}.
\PrintBackRefs{\CurrentBib}

\bibitem [\protect \citeauthoryear {%
Slater%
\ \BBA {} Villarini%
}{%
Slater%
\ \BBA {} Villarini%
}{%
{\protect \APACyear {2016}}%
}]{%
slater2016recent}
\APACinsertmetastar {%
slater2016recent}%
\begin{APACrefauthors}%
Slater, L\BPBI J.%
\BCBT {}\ \BBA {} Villarini, G.%
\end{APACrefauthors}%
\unskip\
\newblock
\APACrefYearMonthDay{2016}{}{}.
\newblock
{\BBOQ}\APACrefatitle {Recent trends in US flood risk} {Recent trends in us flood risk}.{\BBCQ}
\newblock
\APACjournalVolNumPages{Geophysical Research Letters}{43}{24}{12--428}.
\PrintBackRefs{\CurrentBib}

\bibitem [\protect \citeauthoryear {%
TWDB%
}{%
TWDB%
}{%
{\protect \APACyear {2024}}%
}]{%
web_floodquilt}
\APACinsertmetastar {%
web_floodquilt}%
\begin{APACrefauthors}%
TWDB, T\BPBI W\BPBI D\BPBI B.%
\end{APACrefauthors}%
\unskip\
\newblock
\APACrefYearMonthDay{2024}{}{}.
\newblock
\APACrefbtitle {"Flood Quilt (2024)".} {"flood quilt (2024)".}
\newblock
\begin{APACrefURL} \url{https://twdb-flood-planning-resources-twdb.hub.arcgis.com/pages/flood-quilt-2024} \end{APACrefURL}
\PrintBackRefs{\CurrentBib}

\bibitem [\protect \citeauthoryear {%
{'USDA NRCS'}%
}{%
{'USDA NRCS'}%
}{%
{\protect \APACyear {2023}}%
}]{%
usdanrcsGriddedSoilSurvey2023}
\APACinsertmetastar {%
usdanrcsGriddedSoilSurvey2023}%
\begin{APACrefauthors}%
{'USDA NRCS'}.%
\end{APACrefauthors}%
\unskip\
\newblock
\APACrefYearMonthDay{2023}{}{}.
\newblock
\APACrefbtitle {Gridded {{Soil Survey Geographic}} ({{gSSURGO}}) {{Database}} for the {{Conterminous United States}}.} {Gridded {{Soil Survey Geographic}} ({{gSSURGO}}) {{Database}} for the {{Conterminous United States}}.}
\PrintBackRefs{\CurrentBib}

\bibitem [\protect \citeauthoryear {%
Vaswani%
\ \protect \BOthers {.}}{%
Vaswani%
\ \protect \BOthers {.}}{%
{\protect \APACyear {2017}}%
}]{%
vaswani2017attention}
\APACinsertmetastar {%
vaswani2017attention}%
\begin{APACrefauthors}%
Vaswani, A.%
, Shazeer, N.%
, Parmar, N.%
, Uszkoreit, J.%
, Jones, L.%
, Gomez, A\BPBI N.%
\BDBL {}Polosukhin, I.%
\end{APACrefauthors}%
\unskip\
\newblock
\APACrefYearMonthDay{2017}{}{}.
\newblock
{\BBOQ}\APACrefatitle {Attention is all you need} {Attention is all you need}.{\BBCQ}
\newblock
\APACjournalVolNumPages{Advances in neural information processing systems}{30}{}{}.
\PrintBackRefs{\CurrentBib}

\bibitem [\protect \citeauthoryear {%
Vijaymeena%
\ \BBA {} Kavitha%
}{%
Vijaymeena%
\ \BBA {} Kavitha%
}{%
{\protect \APACyear {2016}}%
}]{%
vijaymeena2016survey}
\APACinsertmetastar {%
vijaymeena2016survey}%
\begin{APACrefauthors}%
Vijaymeena, M.%
\BCBT {}\ \BBA {} Kavitha, K.%
\end{APACrefauthors}%
\unskip\
\newblock
\APACrefYearMonthDay{2016}{}{}.
\newblock
{\BBOQ}\APACrefatitle {A survey on similarity measures in text mining} {A survey on similarity measures in text mining}.{\BBCQ}
\newblock
\APACjournalVolNumPages{Machine Learning and Applications: An International Journal}{3}{2}{19--28}.
\PrintBackRefs{\CurrentBib}

\bibitem [\protect \citeauthoryear {%
Wagner%
, Yan%
\BCBL {}\ \BBA {} Yanai%
}{%
Wagner%
\ \protect \BOthers {.}}{%
{\protect \APACyear {2017}}%
}]{%
wagner2017k}
\APACinsertmetastar {%
wagner2017k}%
\begin{APACrefauthors}%
Wagner, F.%
, Yan, Y.%
\BCBL {}\ \BBA {} Yanai, I.%
\end{APACrefauthors}%
\unskip\
\newblock
\APACrefYearMonthDay{2017}{}{}.
\newblock
{\BBOQ}\APACrefatitle {K-nearest neighbor smoothing for high-throughput single-cell RNA-Seq data} {K-nearest neighbor smoothing for high-throughput single-cell rna-seq data}.{\BBCQ}
\newblock
\APACjournalVolNumPages{BioRxiv}{}{}{217737}.
\PrintBackRefs{\CurrentBib}

\bibitem [\protect \citeauthoryear {%
Xu%
, Skoularidou%
, Cuesta-Infante%
\BCBL {}\ \BBA {} Veeramachaneni%
}{%
Xu%
\ \protect \BOthers {.}}{%
{\protect \APACyear {2019}}%
}]{%
ctgan}
\APACinsertmetastar {%
ctgan}%
\begin{APACrefauthors}%
Xu, L.%
, Skoularidou, M.%
, Cuesta-Infante, A.%
\BCBL {}\ \BBA {} Veeramachaneni, K.%
\end{APACrefauthors}%
\unskip\
\newblock
\APACrefYearMonthDay{2019}{}{}.
\newblock
{\BBOQ}\APACrefatitle {Modeling Tabular data using Conditional GAN} {Modeling tabular data using conditional gan}.{\BBCQ}
\newblock
\BIn{} \APACrefbtitle {Advances in Neural Information Processing Systems.} {Advances in neural information processing systems.}
\PrintBackRefs{\CurrentBib}

\bibitem [\protect \citeauthoryear {%
Yildirim%
, Just%
\BCBL {}\ \BBA {} Demir%
}{%
Yildirim%
\ \protect \BOthers {.}}{%
{\protect \APACyear {2022}}%
}]{%
yildirim2022flood}
\APACinsertmetastar {%
yildirim2022flood}%
\begin{APACrefauthors}%
Yildirim, E.%
, Just, C.%
\BCBL {}\ \BBA {} Demir, I.%
\end{APACrefauthors}%
\unskip\
\newblock
\APACrefYearMonthDay{2022}{}{}.
\newblock
{\BBOQ}\APACrefatitle {Flood risk assessment and quantification at the community and property level in the State of Iowa} {Flood risk assessment and quantification at the community and property level in the state of iowa}.{\BBCQ}
\newblock
\APACjournalVolNumPages{International journal of disaster risk reduction}{77}{}{103106}.
\PrintBackRefs{\CurrentBib}

\bibitem [\protect \citeauthoryear {%
Yin%
, Wu%
, Wang%
, Lee%
\BCBL {}\ \BBA {} Wei%
}{%
Yin%
\ \protect \BOthers {.}}{%
{\protect \APACyear {2023}}%
}]{%
yin2023integrated}
\APACinsertmetastar {%
yin2023integrated}%
\begin{APACrefauthors}%
Yin, K.%
, Wu, J.%
, Wang, W.%
, Lee, D\BHBI H.%
\BCBL {}\ \BBA {} Wei, Y.%
\end{APACrefauthors}%
\unskip\
\newblock
\APACrefYearMonthDay{2023}{}{}.
\newblock
{\BBOQ}\APACrefatitle {An integrated resilience assessment model of urban transportation network: A case study of 40 cities in China} {An integrated resilience assessment model of urban transportation network: A case study of 40 cities in china}.{\BBCQ}
\newblock
\APACjournalVolNumPages{Transportation Research Part A: Policy and Practice}{173}{}{103687}.
\PrintBackRefs{\CurrentBib}

\bibitem [\protect \citeauthoryear {%
Zhu%
\ \protect \BOthers {.}}{%
Zhu%
\ \protect \BOthers {.}}{%
{\protect \APACyear {2024}}%
}]{%
ZHU2024101739}
\APACinsertmetastar {%
ZHU2024101739}%
\begin{APACrefauthors}%
Zhu, K.%
, Lai, C.%
, Wang, Z.%
, Zeng, Z.%
, Mao, Z.%
\BCBL {}\ \BBA {} Chen, X.%
\end{APACrefauthors}%
\unskip\
\newblock
\APACrefYearMonthDay{2024}{}{}.
\newblock
{\BBOQ}\APACrefatitle {A novel framework for feature simplification and selection in flood susceptibility assessment based on machine learning} {A novel framework for feature simplification and selection in flood susceptibility assessment based on machine learning}.{\BBCQ}
\newblock
\APACjournalVolNumPages{Journal of Hydrology: Regional Studies}{52}{}{101739}.
\newblock
\begin{APACrefURL} \url{https://www.sciencedirect.com/science/article/pii/S2214581824000879} \end{APACrefURL}
\newblock
\begin{APACrefDOI} \doi{https://doi.org/10.1016/j.ejrh.2024.101739} \end{APACrefDOI}
\PrintBackRefs{\CurrentBib}

\end{thebibliography}
